\documentclass[pdflatex, Numbered, sn-basic]{sn-jnl}

\usepackage{graphicx}%
\usepackage{multirow}%
\usepackage{amsmath,amssymb,amsfonts}%
\usepackage{amsthm}%
\usepackage{mathrsfs}%
\usepackage[title]{appendix}%
\usepackage{xcolor}%
\usepackage{textcomp}%
\usepackage{manyfoot}%
\usepackage{booktabs}%
\usepackage{algorithm}%

\usepackage{anyfontsize}
\usepackage{amstext}
\usepackage{times} 
\usepackage[textsize=tiny]{todonotes}
\hypersetup{hidelinks}
\setuptodonotes{inline}
\usepackage{url}

\usepackage{latexsym}
\usepackage{enumitem}

\usepackage{algorithm}
\usepackage{algorithmic}
\usepackage[caption=false]{subfig}

\newcommand\footnoteref[1]{\protected@xdef\@thefnmark{\ref{#1}}\@footnotemark}
\makeatother

\newcommand{\setplus}{\uplus}

\begin{document}

\title[Surrogate Fitness Metrics for Interpretable Reinforcement Learning]{\centering Surrogate Fitness Metrics for Interpretable Reinforcement Learning\footnotetext{
An earlier version of this work was presented at the International Conference on Evolutionary Computation Theory and Applications (ECTA 2024) \cite{Altmann24-REACT}. 
This article extends our conference paper with a thorough hyperparameter analysis, ablation studies investigating the impact of partial- and fidelity-based fitness functions, as well as a more robust assessment of the generated trajectories in terms of their optimality gap and demonstration fidelities.}}

\author*{\fnm{Philipp} \sur{Altmann}\footnotemark}\email{philipp.altmann@ifi.lmu.de}
\author{\fnm{C\'{e}line} \sur{Davignon}}
\author{\fnm{Maximilian} \sur{Zorn}}
\author{\fnm{Fabian} \sur{Ritz}}
\author{\fnm{Claudia} \sur{Linnhoff-Popien}}
\author{\fnm{Thomas} \sur{Gabor}}

\affil{\orgname{LMU Munich}, \city{Munich}, \country{Germany}}

\abstract{

\textbf{Methods:} We employ an evolutionary optimization framework that perturbs initial states to generate informative and diverse policy demonstrations. A joint surrogate fitness function guides the optimization by combining local diversity, behavioral certainty, and global population diversity. To assess demonstration quality, we apply a set of evaluation metrics, including the reward-based optimality gap, fidelity interquartile means (IQMs), fitness composition analysis, and trajectory visualizations. Hyperparameter sensitivity is also examined to better understand the dynamics of trajectory optimization.

\textbf{Results:} Our findings demonstrate that optimizing trajectory selection via surrogate fitness metrics significantly improves interpretability of RL policies in both discrete and continuous environments. In gridworld domains, evaluations reveal significantly enhanced demonstration fidelities compared to random and ablated baselines. In continuous control, the proposed framework offers valuable insights, particularly for early-stage policies, while fidelity-based optimization proves more effective for mature policies. 

\textbf{Conclusion:} By refining and systematically analyzing surrogate fitness functions, this study advances the interpretability of RL models. The proposed improvements provide deeper insights into RL decision-making, benefiting applications in safety-critical and explainability-focused domains.

Code and videos are available at 
\urlstyle{tt}\url{https://github.com/philippaltmann/REACT}.
}


\maketitle
\clearpage

\section{Introduction}

Interpreting the decision-making process of \textit{reinforcement learning} (RL) policies remains a critical challenge, particularly as RL agents are increasingly deployed in complex, real-world applications. Unlike static supervised learning models, RL policies determine actions based on dynamic interactions with the environment, making their internal reasoning difficult to visualize and analyze. Traditionally, the evaluation of RL policies has focused on their ability to maximize cumulative rewards. However, such an evaluation offers only a limited perspective, as it primarily reflects optimal behavior under training conditions while providing little insight into how a policy generalizes to novel or edge-case scenarios.
\\[2pt]
To enable a more comprehensive understanding of RL models, we propose \textit{Revealing Evolutionary Action Consequence Trajectories} (REACT), a framework that systematically generates diverse, interpretable demonstrations of an RL policy’s behavior. Rather than relying solely on reward-based assessments, REACT employs a surrogate fitness function to evaluate the added interpretability value of different trajectories. By optimizing the initial states through an evolutionary algorithm, REACT generates a pool of trajectories that expose edge-case behaviors and decision-making uncertainties.
\\[2pt]
A key challenge in this approach is the design of an appropriate surrogate fitness effectively capturing the interpretability of a given trajectory. We propose a joint metric that balances local diversity, representing deviations from the expected behavior within a single trajectory, and global diversity, ensuring that the generated set of demonstrations provides broad coverage of the policy’s behavioral space. Additionally, we incorporate action certainty as an indicator of confidence in decision-making, revealing states where the policy exhibits unexpected behavior. This combination of surrogate metrics enables a more structured and systematic approach to policy interpretability, reducing the reliance on human evaluators.
\\[2pt]
To extend our initial investigation, we conduct additional ablation studies to assess the impact of various factors on the interpretability of generated trajectories. Specifically, we analyze the effect of partial reward information and fidelity-based reward formulations on the diversity and informativeness of the demonstrations. Furthermore, we introduce a detailed hyperparameter analysis, systematically evaluating the influence of encoding length, population size, selection mechanisms, and mutation rates on the effectiveness of REACT. Lastly, we provide a quantitative assessment of the optimality gap and fidelity IQM\footnote{For a detailed introduction to the statistical evaluation tools used in this work, including the optimality gap and interquartile mean (IQM) metrics, we refer the reader to the rliable framework \cite{agarwal2021deep}} of the generated trajectories, measuring how closely the extracted demonstrations align with policy behavior under different training conditions.
Overall, our contributions are as follows:\\[-8pt]
\begin{itemize}
\item We propose a joint surrogate fitness function that balances local and global diversity with action certainty to assess the interpretability of RL-generated trajectories. Additionally, we provide ablation studies on the role of partial and fidelity-based trajectory evaluation. 
\item We introduce REACT, an evolutionary optimization framework that selects diverse initial states to generate informative demonstrations of a policy’s behavior, and investigate the impact of the state encoding, population size, selection strategy, and crossover- and mutation rates on the diversity of generated trajectories.
\item We analyze the resulting trajectories in terms of their optimality gap and fidelity, quantifying their effectiveness in capturing meaningful variations in policy behavior.
\end{itemize}

\section{Preliminaries}

\subsection{Reinforcement Learning}

We consider the standard RL framework, where an agent interacts with an environment formalized as a \textit{Markov Decision Process (MDP)}, defined by the tuple $M = \langle\mathcal{S}, \mathcal{A}, \mathcal{P}, \mathcal{R}, \mu, \gamma \rangle$ \citep{puterman1990markov}. Here, $\mathcal{S}$ represents the set of states, $\mathcal{A}$ the set of actions, $\mathcal{P}(s_{t+1} | s_t, a_t)$ the transition probability of reaching state $s_{t+1}$ given action $a_t$ in state $s_t$ at timestep $t$, and $r_t=\mathcal{R}(s_t, a_t, s_{t+1})$ the scalar reward received after transitioning to $s_{t+1}$. The agent starts in an initial state \( s_0 \sim \mu \) and follows a policy \( \pi(a_t | s_t) \), determining action probabilities based on the current state. The objective of RL is to learn an optimal policy \( \pi^* \) that maximizes the expected \textit{discounted return}, defined as:
$$G_t = \sum_{k=0}^{\infty} \gamma^k r_{t+k},$$
where \( \gamma \in [0,1) \) is the discount factor determining the importance of future rewards. \citep{suttonbarto01}
\\[4pt]
Policy-based methods directly approximate the optimal policy by optimizing a \textit{surrogate objective} over trajectories $\tau = \langle s_0, a_0, r_1, \dots, s_t, a_t, r_{t+1}\rangle$, sampled from the policy’s interaction with the environment. Among popular approaches, \textit{Proximal Policy Optimization} (PPO) \citep{schulman2017proximal} refines policy updates by constraining them within a clipped objective to improve stability, while \textit{Soft Actor-Critic} (SAC) \citep{SAC} balances exploration and exploitation using entropy regularization. Both methods have demonstrated versatility across discrete and continuous environments, making them suitable for training policies that serve as subjects for our interpretability studies.
\\[4pt]
Despite their empirical success, deep RL policies remain largely \textit{black-box models}, where the reasoning behind specific decisions is difficult to infer. Even when an RL policy appears optimal, its behavior in unanticipated states or edge cases can be unpredictable, necessitating techniques to extract and analyze such scenarios systematically.

\subsection{Explainability in Reinforcement Learning}

The need for \textit{interpretable RL} has driven research into methods that clarify how agents make decisions. RL interpretability methods can be categorized based on their \textit{scope}, the type of \textit{representation}, and the level of \textit{model accessibility} \cite{Li2022}:
\begin{itemize}
    \item \textbf{Scope:} Some methods provide \textit{local} interpretability by explaining individual state-action decisions, while others focus on \textit{global} interpretability by summarizing general policy behaviors.
    \item \textbf{Representation:} Approaches can leverage \textit{feature importance analysis} \citep{lundberg2017unified}, \textit{model response visualizations} \citep{ribeiro2016should}, or \textit{demonstration-based summaries} \citep{koh2017understanding,pleiss2020identifying} to communicate insights. 
    \item \textbf{Model Accessibility:} Techniques range from \textit{model-agnostic} \citep{pleiss2020identifying, ribeiro2016should} approaches, applicable to any RL agent, to \textit{model-specific} \citep{koh2017understanding} methods that require differentiability or explicit access to internal policy structures.
\end{itemize}
\vspace{4pt}
Our work aligns with \textit{model-agnostic, demonstration-based global interpretability}, where we extract trajectories revealing diverse and unexpected behaviors of a trained policy. Instead of directly approximating the policy with a transparent surrogate model, we analyze its behavior through a set of \textit{diverse action-consequence trajectories} that reveal implicit decision-making patterns.

\subsection{Evolutionary Optimization}

To systematically generate diverse and informative demonstrations, we employ \textit{evolutionary optimization}, a population-based approach that iteratively refines candidate solutions through selection, mutation, and recombination. Given a population $\mathbb{P}$ of individuals, where each individual represents an initial state $s_0$, we optimize $\mathbb{P}$ using an evolutionary process guided by a fitness function.
At each generation, individuals undergo \textit{selection, mutation, and crossover}, defined as follows \cite{fogel2006evolutionary}:
\begin{itemize}
    \item \textbf{Selection}: Chooses individuals from the population based on fitness, often using \textit{tournament selection}, commonly applied with $\sigma$ \cite{miller1995genetic}.
    \item \textbf{Mutation}: Modifies selected individuals by applying a bit-flip operation to their state encoding with probability $p_m$.
    \item \textbf{Crossover}: Combines features from two parents to create new offspring, with probability $p_c$, increasing diversity.
\end{itemize}
Unlike traditional applications that seek a \textit{single best solution}, our method values \textit{population diversity} (similar to \cite{ishibuchi2008evolutionary, neumann2019evolutionary}, e.g., measured by the means of \cite{gabor2018inheritance}) ensuring that the final set of trajectories provides \textit{broad coverage of the policy’s behavioral space}. By leveraging a \textit{joint surrogate fitness function}, we encourage both \textit{local exploration} of individual behaviors and \textit{global diversity} across demonstrations, enabling a structured and interpretable evaluation of RL policies.

\subsection{Surrogate Models for Interpretable Machine Learning}

\textit{Surrogate models} often refer to simplified, computationally efficient approximations of more complex systems, used to analyze, optimize, or interpret their behavior \cite{queipo2005surrogate}. 
These models act as \textit{proxies}, capturing essential characteristics of the original system while reducing computational complexity and are widely used in RL \cite{schulman2017proximal, Altmann24-DIRECT}.
Furthermore, they are widely employed to enhance explainability in machine learning (ML) and reinforcement learning (RL) systems, offering insights that are otherwise difficult to extract from high-dimensional, nonlinear function approximations such as deep neural networks. 
Methods such as \textit{Local Interpretable Model-agnostic Explanations (LIME)} \cite{ribeiro2016should} approximate decision boundaries by fitting simple models in local neighborhoods.
Furthermore, \textit{behavioral surrogates} have bee used to analyze policy behavior, assess robustness, and generate interpretable summaries of decision-making \cite{SEQUEIRA2020103367, puiutta2020explainable}.
Our work builds on these principles by employing a \textit{joint fitness function} as a surrogate measure for the interpretability of RL trajectories. Unlike conventional explainability techniques that analyze feature importance or decision boundaries, we focus on \textit{policy-wide behavioral diversity}, ensuring that selected trajectories reveal nuanced aspects of the agent’s decision-making process.

\clearpage

\section{Trajectory Interpretability Evaluation}

Evaluating RL-generated trajectories based on \textit{interpretability} requires moving beyond traditional reward-centric assessments. Since the ultimate goal is to provide human analysts with meaningful insights into a policy’s inherent behavior, we need a principled way to quantify how diverse, informative, and representative a given trajectory is. To achieve this, we introduce a \textit{joint surrogate fitness function}, which systematically scores trajectories based on their ability to highlight behavioral variations, expose decision-making uncertainty, and cover a broad spectrum of state-action interactions.
\\[4pt]
Instead of relying on human evaluations or qualitative observations, our fitness function enables a structured and scalable approach to trajectory selection. It comprises three key components:
\\[-4pt]
\begin{enumerate}
    \item \textbf{Local Diversity ($\mathcal{D}_l$)} – Measures the variety of states visited within a single trajectory, capturing deviations from standard paths.
    \item \textbf{Global Diversity ($\mathcal{D}_g$)} – Ensures that the selected trajectory adds meaningful variability to a set of previously evaluated demonstrations.
    \item \textbf{Certainty ($\mathcal{C}$)} – Quantifies the confidence of the policy in its decision-making, revealing trajectories where the agent exhibits \textit{uncertainty or inconsistencies}, which are crucial for interpretability.
\end{enumerate}
\vspace{4pt}
Together, these components balance \textit{exploration of novel behaviors} ($\mathcal{D}_l$), \textit{differentiation from redundant examples} ($\mathcal{D}_g$), and \textit{exposure of decision-making stability} ($\mathcal{C}$). Below, we define each metric and discuss how they collectively determine a trajectory’s interpretability.

\subsection{Local Diversity}

An interpretable trajectory should provide a diverse view of how the policy behaves in different states, rather than merely showing optimal action sequences. \textit{Local diversity} ($\mathcal{D}_l$) quantifies how much of the available state space is covered within a single trajectory:
\begin{equation}\label{eq:local_diversity}
\mathcal{D}_l(\tau) = \frac{| \{ s \in \tau \} |}{| P |},
\end{equation}
where $P=\{P_d\;|\;P_d\subset\mathbb{N},\forall d\in 1,\dots, dim\}$, with $|P|=|P_1|\cdot...\cdot |P_{dim}|$ is the $dim$-dimensional position space extracted by $\rho: \mathcal{S} \mapsto P$ from a state $s$.
Yet, this representation might be extended by other important, moving, or task-specific objects like obstacles or targets.
Furthermore, this position-centric formalization allows us to consider the Euclidean distance between states $||s-s'||_2=\sqrt{(\rho(s)-\rho(s')^2}$.
A higher $\mathcal{D}_l$ score indicates that the trajectory explores \textit{a wider range of states}, making it more informative.
\\[4pt]
In \textit{continuous environments}, we generalize this by computing a state density metric based on discretized state representations, where $P$ represents the set of all possible discrete state locations in the environment, ensuring meaningful coverage comparisons.
For use in continuous environments, we suggest applying appropriate discretizations to regularize state similarities.

\subsection{Global Diversity}

While local diversity ensures that individual trajectories are informative, \textit{global diversity} ($\mathcal{D}_g$) prevents redundancy across the entire population of sampled trajectories. This metric ensures that each newly added trajectory meaningfully expands the diversity of the demonstration set $\mathcal{T}$, rather than duplicating existing patterns:
\begin{equation}\label{eq:global_diversity}
\mathcal{D}_g(\tau, \mathcal{T}) = \frac{1}{\lceil P\rceil}\min_{\tau'\in \mathcal{T}\setminus\tau}\delta(\tau,\tau'),
\end{equation}
based on the \textit{maximum state distance} $\lceil P\rceil=\max_{s, s' \in \mathcal{S}}||s-s'||_2$ and the \textit{one-way distance} $\delta$ between trajectories $\tau$ and $\tau'$ \citep{Lin2008OneWD}: 
\begin{equation}
\delta(\tau,\tau') =  \frac{\sum_{s\in\tau} d(s,\tau') + \sum_{s'\in\tau'} d(s',\tau)}{|\tau|+|\tau'|},
\end{equation}
where $d(s',\tau)$ represents the state-to-trajectory distance: $$d(s',\tau)=\min_{s\in\tau}(||s-s'||_2),$$ ensuring that the new trajectory \textit{meaningfully differs} from existing ones. This prevents over-reliance on common behaviors and encourages exposure to \textit{rare, unexpected} policy executions.
\\[4pt]
Ultimately, even if $\mathcal{T}$ contains only optimal yet maximally dissected behavior to reach the target, presenting such diverse demonstrations increases the overall interpretability of $\pi$. 
Note that, even though only defined for disturbing the agent's position, further deviations, such as altering layouts, are formally not precluded. 
However, calculating the global diversity might require using a different distance metric, like the Levenshtein distance, instead.

\subsection{Certainty}

Many interpretability methods overlook the certainty of the policy in its decision-making, which is a crucial factor in assessing model reliability. Low-certainty regions indicate states where the policy is undecided, unstable, or exhibits unexpected shifts in behavior. To capture this, we define \textit{certainty} ($\mathcal{C}$) as:
\begin{equation}\label{eq:certainty}
\mathcal{C}(\tau) = \frac{1}{| \tau |} \sum_{s, a \in \tau} \pi(a | s).
\end{equation}
\\[4pt]
A \textit{low certainty score} highlights trajectories where the policy \textit{hesitates or oscillates} between competing actions. These areas are particularly relevant for interpretability, as they reveal potential weaknesses in the learned policy.

\subsection{Joint Fitness Function}

By combining all three metrics, we define the \textit{joint fitness function} for assessing the interpretability of a trajectory $\tau$ in relation to the set of already explored demonstrations $\mathcal{T}$:

\begin{equation}\label{eq:joint_fitness}
\mathcal{F}(\tau,\mathcal{T}) = \mathcal{D}_g(\tau, \mathcal{T}) + \mathcal{F}_l,
\end{equation}
where
\begin{equation}\label{eq:min-local}
\mathcal{F}_l = \min_{t\in\mathcal{T}}\left|\left|\begin{pmatrix}\mathcal{D}_l(\tau)\\\mathcal{C}_\pi(\tau)\end{pmatrix}-\begin{pmatrix}\mathcal{D}_l(t)\\\mathcal{C}_\pi(t)\end{pmatrix} \right|\right|_2.\end{equation}
\\[4pt]
To reflect the trajectory-specific metrics of local diversity and certainty in relation to the set of all trajectories $\mathcal{T}$ (considered for calculating the \textit{global diversity}), we consult these measures only regarding their minimum distance between $\tau$ and $\mathcal{T}$.
We chose the minimum distance of both local metrics to encourage individuals to maximize their local distance to the closest individual, thereby promoting diverse or uncertain behavior. 
This formulation ensures that \textit{newly selected trajectories balance} the following criteria:
\\[-4pt]
\begin{itemize}
    \item \textbf{Coverage of unique behaviors ($\mathcal{D}_g$)} – Encouraging the discovery of unseen trajectories.
    \item \textbf{Exploration of diverse state-space interactions ($\mathcal{D}_l$)} – Ensuring variety in visited states.
    \item \textbf{Exposure of decision-making inconsistencies ($\mathcal{C}$)} – Highlighting areas of uncertainty or failure.
\end{itemize}
\vspace{4pt}
Importantly, the \textit{ratio of} $\mathcal{D}_l$ \textit{to} $\mathcal{C}$ prioritizes trajectories where diversity is maximized relative to policy uncertainty. This promotes informative yet challenging scenarios that are most useful for interpretability.
\\[4pt]
By introducing a structured, quantitative approach to trajectory evaluation, we move beyond qualitative inspection and subjective assessment. The proposed joint fitness function enables scalable, reproducible, and insightful evaluations of RL policies. 
In \autoref{sec:eval} we will analyze the impact of using different combinations of these surrogate interpretability metrics to optimize a pool of evaluation trajectories.
Additionally, we will compare their performance to simply using the demonstration Fidelity, commonly used to quantitatively assess the interpretability of a set of demonstrations, as the optimization target. 

\clearpage
\section{Revealing Evolutionary Action Consequence Trajectories}

\begin{figure*}[ht]\centering  
  \includegraphics[width=0.7\linewidth]{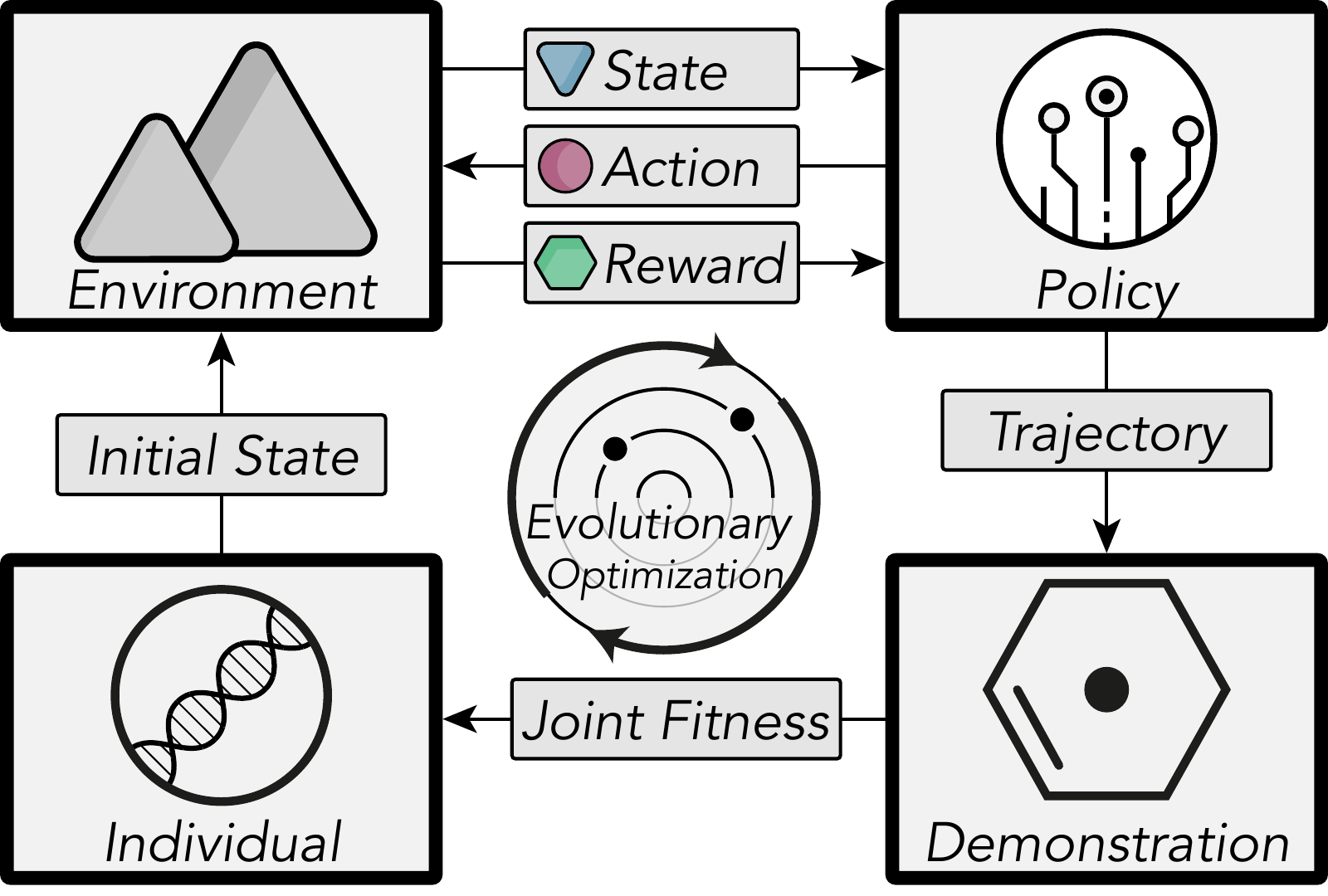}
  \caption{REACT Architecture}\label{fig:react}
\end{figure*}

To optimize a set of demonstrations for interpreting a given policy based on the previously defined fitness criterion, we introduce \textit{Revealing Evolutionary Action Consequence Trajectories} (REACT). This method optimizes a population of initial states that generate diverse demonstrations, facilitating a comprehensive understanding of policy behavior. Rather than merely showcasing the most optimistic optimal behavior, REACT aims to enhance the traceability of learned behaviors, thereby fostering greater trust in the black-box policy model. Unlike conventional evolutionary approaches that focus on identifying a single best-performing individual, our approach considers the entire population.
The overall architecture of REACT is illustrated in Fig.\ref{fig:react} and outlined in Alg.\ref{alg:react}.

\noindent
REACT is designed to iteratively refine a population of initial states to produce informative and diverse trajectories that reveal policy behavior. The process begins by generating an initial population of individuals, each corresponding to a specific initial state. These individuals are evaluated based on the trajectories they generate when following the given policy. Fitness is determined according to a defined criterion that accounts for diversity and informativeness of the demonstrations.\\[4pt]
Subsequently, the algorithm applies evolutionary operations, including selection, mutation, and crossover, to generate new individuals. Selected individuals undergo recombination to introduce variation, while mutations further diversify the population. Through multiple generations, REACT refines the population, promoting the selection of initial states that yield the most revealing demonstrations. The process continues until a predefined number of generations is reached, ultimately producing a set of optimized trajectories.\\[4pt]

\begin{algorithm*}[t]
\caption{Revealing Evolutionary Action Consequence Trajectories (REACT)\cite{Altmann24-REACT}\protect\footnotemark}\label{alg:react}
\begin{algorithmic}[1]
\REQUIRE{$\mathcal{P},\mu,\pi$ \hspace{90pt}\textit{$\triangleright$ We use a policy trained with a single initial state}}
\STATE{$\mathbb{P} \gets \langle s_0 \sim \mu\rangle_p$ ; $\mathcal{T} \gets \emptyset$ \hspace{45.5pt}\textit{$\triangleright$ Generate initial population of size $p$ and empty $\mathcal{T}$}}
\FOR{individual $\mathcal{I} \in \mathbb{P}$}\label{alg:ln:Fs}
\STATE{$\tau \sim \mathcal{P}_{\pi, s_0}$\hspace{94pt} \textit{$\triangleright$ Sample trajectory $\tau_\mathcal{I}$ from initial state $s_0$}}
\STATE{$\mathcal{F}_\mathcal{I}(\tau,\mathcal{T})$\hspace{100pt}\textit{$\triangleright$ Calculate Fitness of $\mathcal{I}$ w.r.t. to phenotype $\tau$ and \\ \hspace{141pt} previous demonstrations $\mathcal{T}$ according to Eq.~\eqref{eq:joint_fitness}}}
\STATE{$\mathcal{T}\gets\mathcal{T}\cup\tau$\hspace{87.5pt} \textit{$\triangleright$ Update demonstrations $\mathcal{T}$}}
\ENDFOR\label{alg:ln:Fe}
\FORALL{generations $g$}
\STATE{$\mathbb{O}\gets \texttt{mutants}_{p_m}(\mathbb{P})\setplus\texttt{children}_{p_c,\mathcal{F}}(\mathbb{P})$\hspace{4pt}$\triangleright$ \textit{Generate offspring from mutation and \\\hspace{136pt}crossover with $p_m$, $p_c$, $\mathcal{F}_\mathcal{I}$, and tournament selection}}
\FOR{individual $\mathcal{I} \in \mathbb{O}$}
\STATE{$\tau \sim \mathcal{P}_{\pi, s_0}$\hspace{84pt} \textit{$\triangleright$ Sample trajectory $\tau_\mathcal{I}$ from initial state $s_0$}}
\STATE{$\mathcal{F}_\mathcal{I}(\tau,\mathcal{T})$\hspace{90pt}\textit{$\triangleright$ Calculate Fitness according to Eq.~\eqref{eq:joint_fitness}}}
\STATE{$\mathcal{T}\gets\mathcal{T}\cup\tau$\hspace{80pt}\textit{$\triangleright$ Update demonstrations $\mathcal{T}$}}
\ENDFOR
\STATE{$\mathbb{P}\gets\texttt{migration}(\mathbb{P}\setplus\mathbb{O}, \mathcal{F}, p)$\hspace{8pt}\textit{$\triangleright$ Select $p$ best individuals for next generation from \\ \hspace{141pt} population and offspring according to their fitness}}
\STATE{$\mathcal{T} \gets \mathcal{T} \setminus \{\tau_{\mathcal{I}} \; \vert \; \mathcal{I} \notin \mathbb{P}\}$\hspace{45pt}\textit{$\triangleright$ Remove extinct demonstrations}}
\ENDFOR
\RETURN $\mathcal{T}$
\end{algorithmic}
\end{algorithm*}

\footnotetext{\label{note:appendix} All required implementations, appendices, and video renderings are available at \url{https://github.com/philippaltmann/REACT}.}

\subsection{State Encoding}\label{sec:state_encoding}

In the following, we elaborate on our encoding design of the initial state $s_0$, defining the individual's genotype. The initial state is determined by the agent's initial position. In most environments, the initial state is either fixed or randomly chosen (sampling from $\mu$), as is the goal state. Fixed and randomly selected initial states and goals have a significant impact on the behavior a trained agent will exhibit. Thus, when configuring the environment, it is crucial to account for initial states and goals.
By controlling these factors, we can reduce randomness in the agent's behavior, leading to more consistent evaluations. Furthermore, this approach allows us to explore unlikely initial states, enabling observations of the agent's responses to rare or atypical scenarios that would not commonly occur during normal episodes.
\\[4pt]
To represent variations in the initial state, we employ a binary encoding scheme, which is particularly advantageous due to its compatibility with recombination and mutation operators. However, a key challenge with binary encoding is that with $m$ bits, exactly $2^m$ possible states can be represented. For instance, in a $9\times9$ gridworld environment with 81 potential initial positions, a 6-bit encoding would only allow for 64 states, while a 7-bit encoding could accommodate 128 states. To address this, we apply inverse normalization.
Every dimension $d$ of the startup state space is encoded using a binary encoding $e$ of length $m$.
Each segment of the binary encoding is first mapped to an integer, then normalized, and subsequently mapped to the valid state space of that dimension. The resulting values form the final initial state for the environment.
\\[4pt]
This mapping enables the representation of discrete state spaces of varying sizes, as well as continuous state spaces with a defined precision. The normalization and inverse normalization steps differ slightly between discrete and continuous state spaces.

\subsubsection*{Discrete State Space}
For a discrete state space, the state encoding is divided into $\vert d \vert$ equally sized segments $e_d$, which are processed as follows:
The binary encoding $e_d$ of length $m$ is first converted to an integer value, which is then normalized by dividing it by the number of possible integer values for that encoding length:
$$normalizedState = \frac{int(e_d)}{2^m} $$
The resulting $normalizedState$ lies within $[0,1[$. To map this normalized value to a valid state in the given dimension, we use the minimum and maximum state values, performing inverse normalization as follows:
$$state = \lfloor normalizedState * (maxValue +1 - minValue) + minValue \rfloor$$
Since the number of encoded states often exceeds the number of actual states in the environment, certain states may occur with slightly different probabilities. This probability discrepancy can be minimized by increasing the encoding length.
For instance, in a $9\times9$ gridworld, each dimension (row and column) requires a binary encoding of at least 4 bits to represent the 9 possible states ($maxValue = 8$, $minValue = 0$).

\subsubsection*{Continuous State Space}
For a continuous state space, integer values are not required, necessitating a slightly different approach. The encoding length is chosen based on the desired precision. The normalization step is given by:
$$normalizedState = \frac{int(e_d)}{2^m - 1},$$ 
where the $normalizedState$ now falls within $[0,1]$. The inverse normalization is then applied without rounding:
$$state = normalizedState * (maxValue - minValue) + minValue$$
For continuous state spaces, increasing the encoding length improves precision, with the smallest possible state value difference determined by:
$$stateValueDistance = \frac{maxValue - minValue}{2^m - 1}$$
\\[4pt]
The impact of different encoding lengths will be further analyzed in \autoref{sec:hyperparameters}.

\clearpage
\subsection{Fitness Evaluation}

The fitness of each individual is assessed by sampling trajectories $\tau$ from the environment, starting from the respective initial state and following policy $\pi$. To enhance comparability, duplicate consecutive states within $\tau$ are removed. These demonstrations represent the individual's phenotype, directly influencing its fitness as a representative of the given policy model.
\\[4pt]
Individual fitness is computed based on Eq.~\eqref{eq:joint_fitness}, considering both the trajectory $\tau$ and the existing demonstration set $\mathcal{T}$, ensuring that each individual's performance is evaluated in the context of the entire demonstration pool. While REACT samples experiences from the environment, it does not aim to improve the given policy. However, its architecture could be extended to function as an automated adversarial curriculum for generating scenarios to enhance further policy training.

\subsection{Evolutionary Process}

Following the evaluation of the initial generation, top-performing individuals are selected via tournament selection to generate new individuals through recombination. The recombination operator is applied with a predefined probability $p_c \in [0,1]$, using single-point crossover to produce offspring. The newly generated individuals are then incorporated into the population.
\\[4pt]
A mutation operator, executed with probability $p_m \in [0,1]$, applies a single-bit flip to a randomly chosen bit in an individual’s encoding. Since our approach prioritizes population diversity, the original individual is retained alongside its mutated counterpart, ensuring an elitist evolutionary strategy.
\\[4pt]
Once offspring are evaluated as described above, the population is pruned to maintain the intended size $p$ by removing individuals with the lowest fitness scores along with their associated demonstrations. This evolutionary process is repeated for a fixed number of $g$ generations, iteratively refining the demonstration pool.

\subsection{Hyperparameters}
The effectiveness of REACT is influenced by several hyperparameters that govern the evolutionary process. The population size $p$ determines the number of initial states considered in each generation, impacting the diversity of trajectories. To suit human needs, $p$ should be comprehensibly small and sufficiently diverse \citep{behrens2023fatigue}.The mutation rate $p_m$ controls the probability of introducing random changes to an individual's encoding, while the crossover rate $p_c$ specifies the likelihood of recombining two individuals to create offspring. Additionally, the number of generations $g$ dictates the extent of evolutionary refinement, balancing computational efficiency and solution quality. If not stated otherwise, we optimize the population of demonstrations over 40 iterations (generations). 
Yet, appropriate tuning of the remaining hyperparameters is crucial for optimizing the interpretability of the resulting demonstrations. 
Therefore, \autoref{sec:hyperparameters} covers a hyperparameter study comparing different values for these parameters.
\clearpage

\section{Related Work}

Understanding and explaining reinforcement learning policies has been an active area of research, intersecting multiple fields including \textit{evolutionary RL}, \textit{robust RL}, \textit{explainable RL}, and \textit{policy testing}. Below, we position our work within these broader contexts.

\subsection*{\textbf{Evolutionary RL}}

Evolutionary approaches have been extensively used to optimize a population of policies \citep{DBLP:journals/corr/abs-1805-07917}, leveraging diversity-driven selection mechanisms to enhance exploration and robustness. Both task-agnostic \citep{DBLP:populationbasedRL} and task-specific \citep{wu2023qualitysimilar} diversity measures have been shown to improve policy performance and adaptability. However, in contrast to these works, we do not aim to improve policies but rather to generate diverse evaluation scenarios that best describe the learned policy within a given environment.  
\\[4pt]
The field of \textit{quality diversity} (QD) optimization has also emerged from considering behavioral novelty as an optimization criterion \citep{lehman2011abandoning}. Such approaches have been used in RL to optimize for \textit{novel and high-performing} behaviors rather than just maximizing rewards. \citet{gabor2019surrogate} explored surrogate-assisted genetic algorithms to improve recommender systems, while \citet{bhatt2022deep} integrated QD into automated environment generation via a surrogate model to improve policy robustness.  
\\[4pt]
We take a similar approach in using an evolutionary optimization framework but with a different objective: improving the interpretability of the learned policy by optimizing a set of diverse policy demonstrations. Unlike novelty search methods that prioritize behavioral distance from the current population, we introduce a \textit{joint fitness function} that combines both \textit{local and global} interpretability criteria to systematically select representative trajectories.  

\subsection*{\textbf{Robust RL}}
Robust reinforcement learning aims to improve generalization by training policies to handle \textit{out-of-distribution} scenarios, such as unexpected initial states or environmental variations. We consider a framework where a policy is trained using a single deterministic initial state and evaluated under varying initializations to expose its \textit{edge-case behavior}. This allows us to assess the learned strategy in previously unseen conditions, revealing implicit policy weaknesses.  
\\[4pt]
Robust RL often seeks to improve generalization capabilities using diverse training configurations \citep{cobbe2020leveraging}, optimized training scenarios \citep{altmann2023crop}, or evolving curricula \citep{parker2022evolving}. In contrast, we focus on \textit{post-hoc} policy analysis rather than improving training dynamics. The generated trajectory representations can, however, be integrated into the training process as \textit{adversarial samples} to refine policy robustness, similar to the adversarial scenario generation approach of \citet{gabor2019scenario}.  
\clearpage

\subsection*{\textbf{Explainable RL}}

Several approaches have been proposed for \textit{interpretable and explainable RL} (XRL), surveyed comprehensively by \citet{HEUILLET2021106685} and \citet{RLinterpretsurvey}. XRL methods can be categorized based on:  
\\[-8pt]
\begin{itemize}
    \item \textbf{Scope:} Local interpretability (explaining specific decisions) vs. global interpretability (summarizing policy behavior).
    \item \textbf{Model Access:} Post-hoc analysis of trained policies \citep{DBLP:policysummarization} vs. intrinsic methods that constrain models to be inherently interpretable \citep{DBLP:HuangHAD17, guo2021edge}.
    \item \textbf{Explanation Type:} Text-based explanations, visual state summaries, rule-based descriptions, or collections of state-action pairs.
\end{itemize}
\vspace{4pt}
Our approach falls into the category of \textit{post-hoc, global} XRL. We generate a collection of demonstration trajectories that illustrate diverse behaviors of a trained policy interacting with its environment. Unlike feature-based attribution methods, REACT extracts structured behavioral summaries by selecting trajectories that maximize interpretability.  
\\[4pt]
\citet{Amir2018HIGHLIGHTSSA} proposed policy summaries consisting of key \textit{critical states}, identified based on their influence on cumulative reward. Similarly, \citet{DBLP:crtiticalstates} introduced the concept of \textit{critical states} in entropy-based RL, where small action deviations lead to large outcome differences. In contrast to these approaches, we do not use value functions or reward gradients to define interpretability. Instead, we optimize a trajectory set using an explicit interpretability-driven fitness function.  
\\[4pt]
Likewise, \citet{SEQUEIRA2020103367} generated \textit{interestingness}-based video summaries by selecting highlights from an agent’s interaction with its environment, focusing on action frequency, certainty, and transition impact. While sharing a similar motivation, our method differs by optimizing \textit{entire} trajectories rather than assembling fragmented high-impact segments. This ensures that our results represent complete behavioral patterns rather than isolated critical moments.  
\\[4pt]
To measure the quality of the selected demonstrations, we adopt the fidelity metric introduced by \citet{guo2021edge}, adapting it to increase fidelity with higher scores. 
We also use this metric as an alternative fitness metric to benchmark the proposed joint fitness.  

\subsection*{\textbf{RL Testing and Policy Verification}}

Similar to REACT, \citet{tappler2022search} used genetic algorithms to search for \textit{interesting traces} in RL policies, while \citet{zolfagharian2023search} optimized full episodes to identify faulty behaviors and train predictive failure models. Unlike these works, which focus primarily on detecting \textit{failure cases}, REACT optimizes the initial state to generate a diverse range of trajectories that provide a balanced representation of policy behavior.  
\\[4pt]
\citet{pang2022mdpfuzz} introduced an RL fuzz-testing framework that perturbs initial conditions to explore policy sensitivities. However, their approach estimates model sensitivity heuristically, whereas REACT systematically applies evolutionary optimization to refine trajectory selection. While RL testing methods generally prioritize failure detection, REACT aims to construct a holistic policy evaluation by capturing both expected and edge-case behaviors.

\section{Evaluation}\label{sec:eval}

\subsection{Setup}

To validate the proposed architecture, we use a simple, fully observable discrete \textit{FlatGrid11} environment with $11\times11$ fields, shown in Fig.~\ref{fig:FlatGrid11} \citep{Altmann_hyphi_gym}. The objective of the policy is to reach the target state, which provides a reward of $+50$. There are no obstacles or holes in this environment that could obstruct the agent’s movement. To encourage efficient navigation, each step incurs a cost of $-1$, ensuring that the optimal policy minimizes the trajectory length. An episode terminates upon reaching the target or after 100 steps.
\\[4pt]
We train a policy using Proximal Policy Optimization (PPO) \citep{schulman2017proximal}, employing the default parameters provided by Stable-Baselines3 \citep{stable-baselines3}. To ensure diverse and suboptimal behavior, training is intentionally stopped early, after 35k steps, just as the agent begins to reliably reach the target. Using an imperfect policy increases the probability that the agent has not yet fully explored the environment, making its decision-making process more interpretable. Additionally, this setup ensures that the policy exhibits both successful and unsuccessful behaviors, allowing REACT to extract diverse and informative demonstrations. It is important to note that the policy is trained from a single initial state, as depicted in Fig.~\ref{fig:FlatGrid11}, making it particularly relevant to investigate the generalization and behavioral diversity when evaluated from different starting conditions.
\\[4pt]
All reported results are averaged over ten random seeds to ensure statistical significance. Each optimization run refines the demonstration set based on a single, pre-trained policy, ensuring that REACT's effects on trajectory diversity and fidelity are robust across multiple initial conditions.
Since trajectory diversity is inherently subjective, we introduce quantitative metrics to evaluate their \textit{behavioral spread}. We adopt two primary statistical measures:  
\\[4pt]
\textbf{Fidelity Interquartile Mean (IQM)}: Fidelity serves as a measure of how well the generated demonstrations approximate and cover the agent's decision-making space. We compute fidelity using the weighted deviation of cumulative rewards across trajectories, adapted from \citep{guo2021edge}:  
\begin{equation}
S = \sum_{\tau \in \mathcal{T}} \frac{|\tau|}{|\mathcal{T}|} \left| \bar{R} - r_\tau \right|,
\end{equation}
where the absolute mean reward is defined as: $\bar{R} = \frac{1}{|\mathcal{T}|} \sum_{\tau \in \mathcal{T}} |r_\tau|,$ and the total trajectory length is given by: $|\mathcal{T}| = \sum_{\tau \in \mathcal{T}} |\tau|$.
Intuitively, fidelity measures the explanatory power of a trajectory set by quantifying its coverage of policy behavior. A higher fidelity score indicates that the selected demonstrations provide a more representative summary of the agent's decision-making process. Following \textit{rliable} \citep{agarwal2021deep}, we report \textit{interquartile means} (IQM), less sensitive to outliers, providing a robust measure of overall demonstration quality.
\\[4pt]
\textbf{Reward Optimality Gap:} The optimality gap quantifies the difference in cumulative reward between the selected demonstrations and the theoretically optimal policy. We compute the optimality gap using rliable's distributional statistics \citep{agarwal2021deep}, which assess how far generated demonstrations deviate from the optimal behavior in terms of expected returns. Instead of merely reporting the mean, we analyze the range and spread of the optimality gap, as a larger range combined with a higher mean value indicates greater behavioral diversity while still capturing high-performing trajectories.  
\\[4pt]
\textbf{State Coverage Heatmaps:} To provide an intuitive overview of the state coverage in the generated demonstrations $\mathcal{T}$, we employ log-scaled state visitation heatmaps. These visualizations depict the frequency with which each state is visited across all demonstrations. Using a logarithmic scale helps regularize the difference between rarely and frequently visited states, making it easier to interpret state coverage. The rationale behind this approach is that visiting a previously unseen state once is highly informative, whereas the difference between visiting a frequently visited state 99 or 100 times is less relevant from an interpretability perspective. These heatmaps provide insights into how REACT enhances trajectory diversity by encouraging exploration of previously underrepresented states.
\\[4pt]
\textbf{Return and Length Distributions:} 
To further characterize the generated demonstrations, we also analyze the final return and trajectory length distributions. While these metrics are crucial in training an optimal policy (maximizing return and minimizing path length), they are not directly optimized by REACT. Instead of focusing on their absolute values, we assess their range and uniformity across individuals. To visualize this, we use box plots, where a larger range between whiskers indicates higher diversity, and a wider interquartile range (IQR) suggests a more even distribution of different behaviors. This analysis ensures that REACT not only selects highly diverse trajectories but also maintains a well-distributed set of policy behaviors.
\\[4pt]
Overall, these evaluation metrics provide a comprehensive framework for assessing the quality and interpretability of the generated demonstrations. By combining fidelity IQM, optimality gap analysis, trajectory statistics, and log-scaled state visitation heatmaps, we ensure that REACT not only captures meaningful variations in policy behavior but also outperforms naive approaches in generating diverse and interpretable demonstrations.
\\[4pt]
\textbf{Baselines:} 
To assess the effectiveness of REACT in generating diverse and interpretable demonstrations, we compare it against several baselines. 
To reflect a \textit{training} validation, we evaluate the policy in the same fixed starting condition used during training. This provides a direct reference for assessing the limitations of evaluating policies in the unaltered training environment. 
Additionally, we compare REACT to a \textit{random search baseline}, implemented as the initial population \(\mathbb{P}_0\) before applying the evolutionary process. This \textit{Random} approach serves as a non-optimized reference, similar to traditional interpretability methods that alter the environment without optimization. By ensuring comparability to REACT while lacking structured trajectory selection, this baseline highlights the added value of evolutionary optimization in refining diverse and representative demonstrations.
Lastly, we include a \textit{fidelity-optimized baseline}, where initial states are explicitly selected by maximizing the fidelity metric. This allows us to determine whether fidelity alone is a sufficient optimization target for generating informative policy demonstrations.  
\\[4pt]
\textbf{Ablations:} 
To further analyze the contribution of different components in REACT’s fitness function, we conduct an \textit{ablation study} by evaluating alternative fitness formulations. These include optimizing trajectories based on only \textit{certainty} (\(\mathcal{C}\)), only \textit{local diversity} (\(\mathcal{D}_l\)), or only \textit{global diversity} (\(\mathcal{D}_g\)). Additionally, we compare REACT’s structured joint fitness function against a \textit{simple sum} of the three components, which does not include the minimum-distance formulation for local diversity. This study helps identify the most important factors for optimizing demonstration diversity and interpretability.

\clearpage
\subsection{Hyperparameters}\label{sec:hyperparameters}

The effectiveness of the REACT framework depends on several key hyperparameters that influence the diversity and interpretability of the generated trajectories. To systematically assess their impact, we evaluate the role of encoding length, which determines the resolution of the initial state representation, and population size, which affects the diversity and convergence of the evolutionary search. Additionally, we analyze the effect of the number of generations on trajectory optimization and study the influence of crossover and mutation probabilities, which control the balance between exploration and exploitation in the evolutionary process. By varying these parameters, we aim to quantify their effect on trajectory diversity, optimality gap, and fidelity, providing insights into their importance for maximizing interpretability.

\subsubsection*{Encoding Length}
\begin{figure}[ht]\centering
\includegraphics[width=0.9\linewidth]{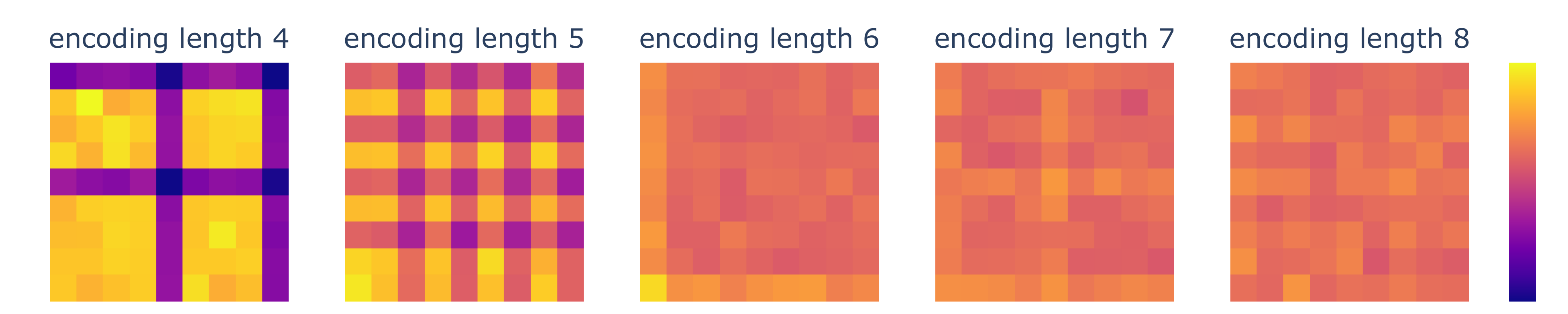}
\caption{Comparing encoding lenghts $n \in [4, 5, 6, 7, 8]$ for encoding random 11x11 states}\label{fig:enc_length_comparison}
\end{figure}

We already approached the topic of choosing the right encoding length in \autoref{sec:state_encoding}.
The encoding of each dimension has to have a sufficient length such that every state in the state space can be represented by it.
However, in an environment with a discrete state space, some startup states could be represented by more encodings than other startup states. 
With our approach, this leads to a higher probability of occurring. 
The difference in probabilities depends on the length of the state encoding. 
Considering a grid size of $11\times11$, the possible startup state are only represented by a grid size of $9\times9$. 
We did not consider cells on the surrounding wall as possible startup states. Each of the two dimensions has 9 possible values, that need to be encoded by a binary encoding with at least 4 bits.
We concluded a simple experiment, to be able to better recommend an encoding length.
We compared five different encoding lengths per dimension: $n \in [4, 5, 6, 7, 8]$.
For each encoding length, we created 81000 random bit encodings of length $2*n$. We transformed them to states in the environment using the method described in \autoref{sec:state_encoding} and plotted them in a heatmap to be able to show the difference in the occurrence of states.
Looking at Figure \ref{fig:enc_length_comparison}, we can see a significant difference in the length of encodings.
Especially for an encoding length of $4$ or $5$, the difference in occurrence is clearly visible. The darker cells are less probable to occur then the lighter cells. This supports the computation of the difference in probability described in \autoref{sec:state_encoding}. For an encoding length of $4$, some states are $2$ times as probable as others. For an encoding length of $5$, this is reduced to $1.33$ times. Looking at the heatmaps and the calculation of the difference in probability, we recommend to use a state encoding length, where a state is less than $1.25$ more likely than any other state. In \texttt{FlatGrid11}, this is the case for an encoding length of $6$. There only exist a few states that are $1.14$ more likely than other states. This difference in probability can be regarded as insignificant for our algorithm.

\subsubsection*{Population Size and Number of Generations}

The population size determines how many demonstrations are shown to the user. On one hand, it is essential to display enough diverse behaviors for the user to gain a comprehensive understanding of the agent’s strategy. On the other hand, executing the genetic algorithm with a large population size while displaying only the top individuals has several drawbacks.
First, a larger population increases the likelihood of retaining individuals with low fitness values, including those that are not meaningfully diverse. This is because survival in the population becomes easier when there are more individuals, reducing the competitive selection pressure. Second, larger population sizes make it more challenging for new individuals to be considered sufficiently diverse, as they must distinguish themselves from a greater number of already evaluated individuals. Generally, the more individuals a candidate must compete against, the less likely it is to be considered novel.
\begin{figure}[ht]\centering\vspace{-16pt}
    \subfloat[All individuals\label{subfig:popsize_iter_a}]{\includegraphics[width=0.5\linewidth]{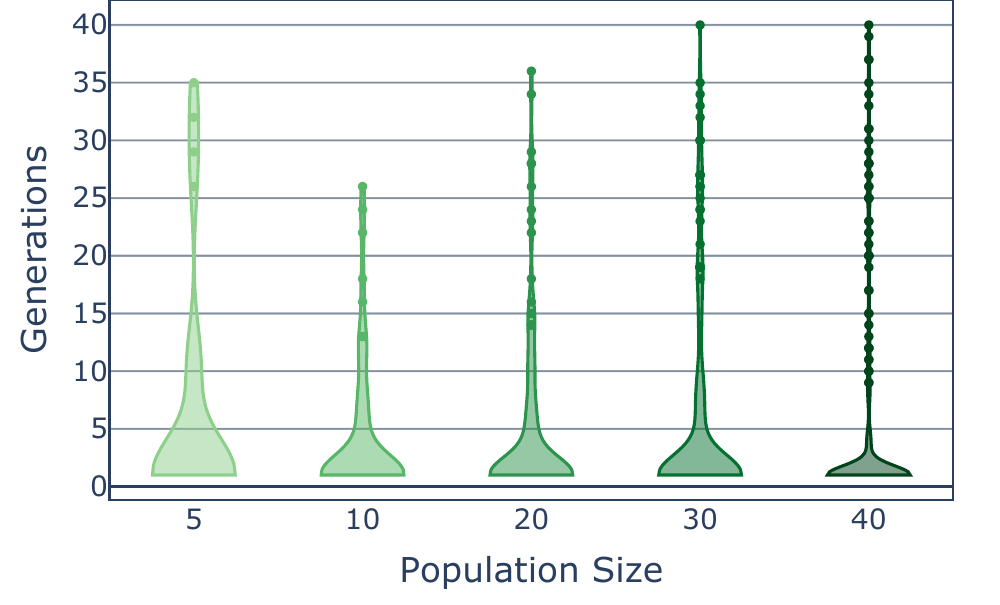}}
    \subfloat[Best 10 individuals\label{subfig:popsize_iter_b}]{\includegraphics[width=0.5\linewidth]{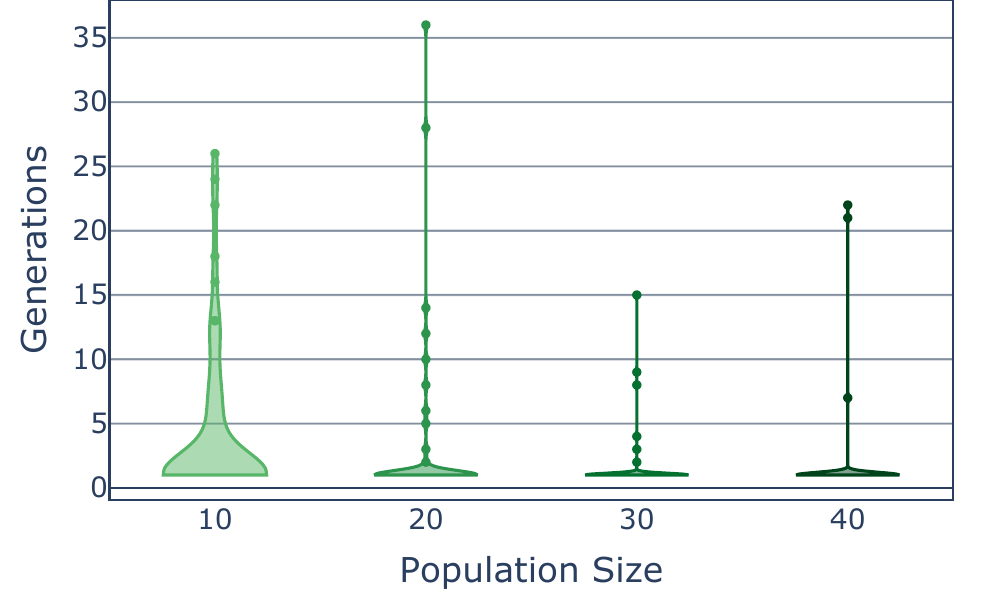}}
    \caption{Comparing individual survival generations over various population sizes.
    }
    \label{fig:popsize_iter}
\end{figure}
\\[4pt]
Figure \ref{fig:popsize_iter} supports this reasoning, illustrating the number of newly added individuals that survived at least one generation for different population sizes, averaged over four random seeds. In Figure \ref{subfig:popsize_iter_a}, which considers all individuals in the population, we observe that larger populations tend to retain more individuals for longer periods. This suggests that with a larger population, new individuals have a higher chance of surviving to the next generation, reducing selection pressure and leading to slower convergence of the evolutionary search. 
In environments with a limited number of possible initial states, this effect becomes even more pronounced. Since diversity is assessed relative to existing individuals, the initial population members have a competitive advantage over newer ones. Early individuals are evaluated against a relatively small set of existing candidates, allowing them to achieve high diversity scores. However, as the algorithm progresses, newer individuals must compete with an increasingly diverse set of previously evaluated individuals, making it more difficult for them to be considered novel.
\\[4pt]
To address this, we restrict the displayed individuals to only the top 10 most diverse ones, as shown in Figure \ref{subfig:popsize_iter_b}. While larger populations allow for more new individuals to enter the population over time, the trend changes when considering only the best-ranked individuals. Here, we observe that increasing the population size leads to earlier algorithm convergence, with fewer new individuals being ranked among the best at later iterations. This suggests that for larger populations, the search space becomes saturated more quickly, limiting the discovery of new, high-quality individuals.
Interestingly, our findings indicate that diversity across different population sizes remains relatively stable, with smaller populations often yielding slightly better results. In particular, larger population sizes tend to retain more individuals with low fitness. Additionally, increasing the population size leads to a slight decrease in diversity.
\\[4pt]
Given that larger populations do not provide significant advantages in terms of diversity while also increasing computational costs, we select a population size of 10 for the subsequent experiments. This choice allows us to balance behavioral diversity while minimizing redundant episodes for the user to review. 
Regarding the number of generations, for a population size of 10, we observe that the algorithm typically converges within 20 iterations, after which only a few new individuals are introduced. Therefore, to ensure sufficient exploration while maintaining computational efficiency, we set the maximum number of iterations to 40.

\subsubsection*{Crossover and Mutation Probability}
Next, we consider the influence of the central hyperparameters of the genetic algorith, the crossover probability and the mutation probability. A higher crossover probability usually leads to more exploitation of individuals with a high fitness value, while a high mutation probability explores the search space to avoid getting stuck in a local optimum. By applying mutation, new diverse individuals are evaluated, that could lead to new insights.
We explored several different settings for the two probabilities: a high crossover probability of $0.9$ with a low mutation probability of $0.25$, a slightly lower crossover probability of $0.75$ with a higher mutation probability of $0.5$, and both of these settings with switched probabilities. Additionally, we added the setting with a crossover probability of $0.9$ and a mutation probability of $0.4$, which we used in the previous experiments.
With this experiment, we mean to investigate, if the search space requires more exploitation of the already evaluated individuals or more exploration of the search space.
\begin{figure}[ht]\centering\vspace{-16pt}
    \subfloat[distribution of the returns in the last generation\label{subfig:prob_reward}]{\includegraphics[width=0.5\linewidth]{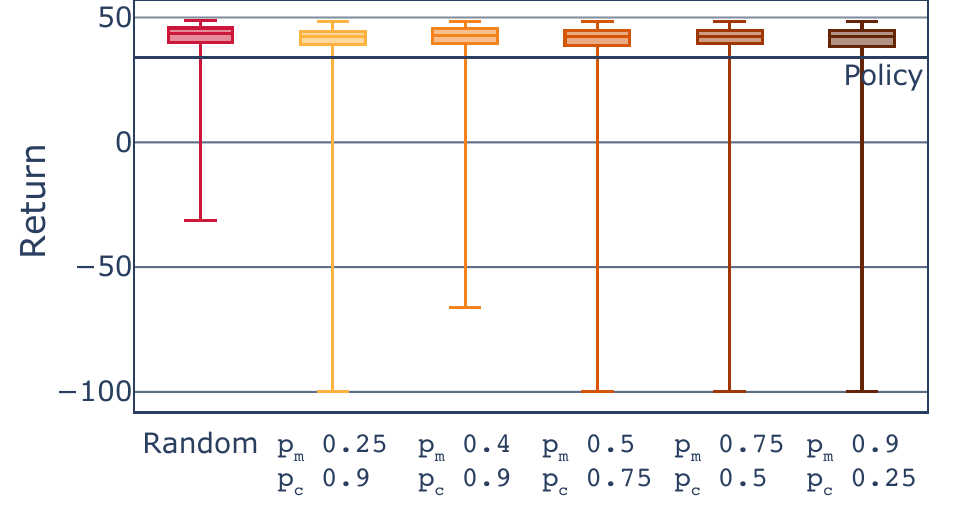}}
    \subfloat[distribution of the trajectory lengths in the last generation\label{subfig:prob_traj_len}]{\includegraphics[width=0.5\linewidth]{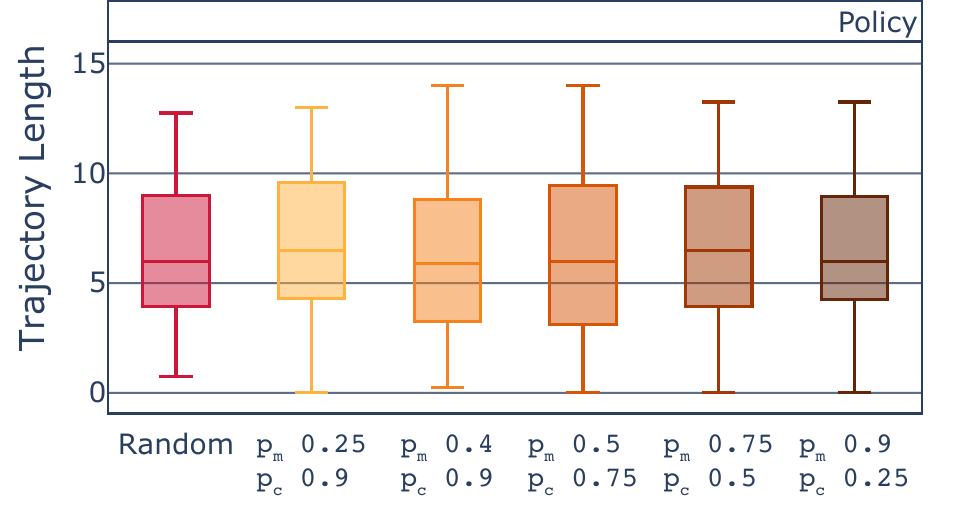}}
    \caption{Diversity of the individuals of the last generation compared by different operator probabilities}
    \label{fig:prob_results}
\end{figure}
\\[4pt]
The results in Figure \ref{fig:prob_results} show, that a low crossover probability paired with a high mutation probability does not lead to significantly better results in terms of the distribution of trajectory lengths than random search. However, looking at the distribution of returns, we can state, that our algorithm shows more diverse behavior than random search. Random search does not always manage to find the startup states that lead to a very interesting behavior, namely the agent's standstill. 
Thus, for the following, we use REACT with a crossover probability of $0.75$ and a mutation probability of $0.5$, offering a tradeoff between mutation and crossover.

\clearpage

\subsection{Gridworld Evaluation}
\begin{figure*}[ht]\centering\vspace{-10pt}
\subfloat[\texttt{FlatGrid11}~\cite{Altmann_hyphi_gym}\label{fig:FlatGrid11}]{\includegraphics[width=0.32\linewidth]{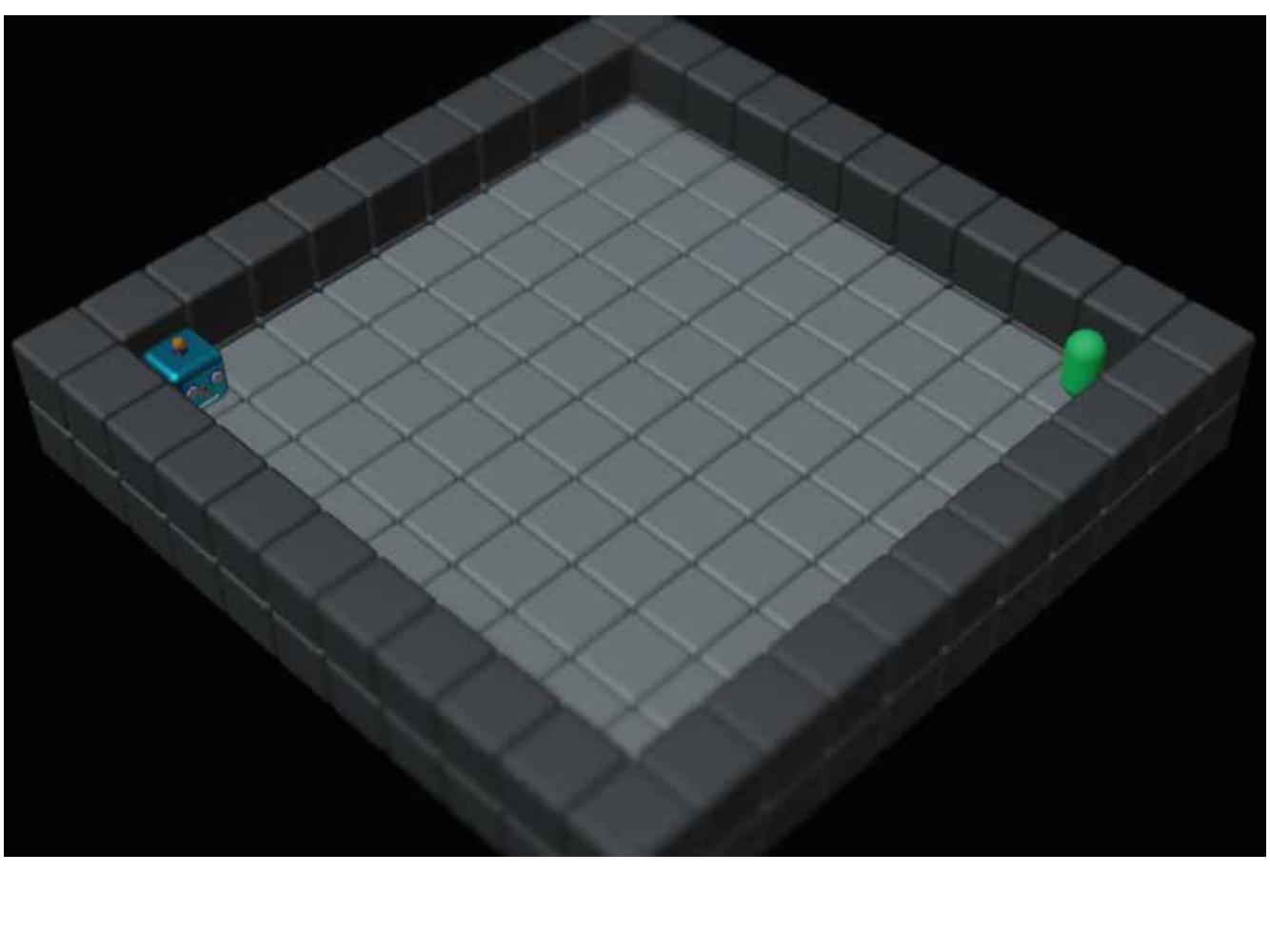}}
\subfloat[Fidelity Optimization\label{fig:FlatGrid-Fidelity}]{\includegraphics[width=0.32\linewidth]{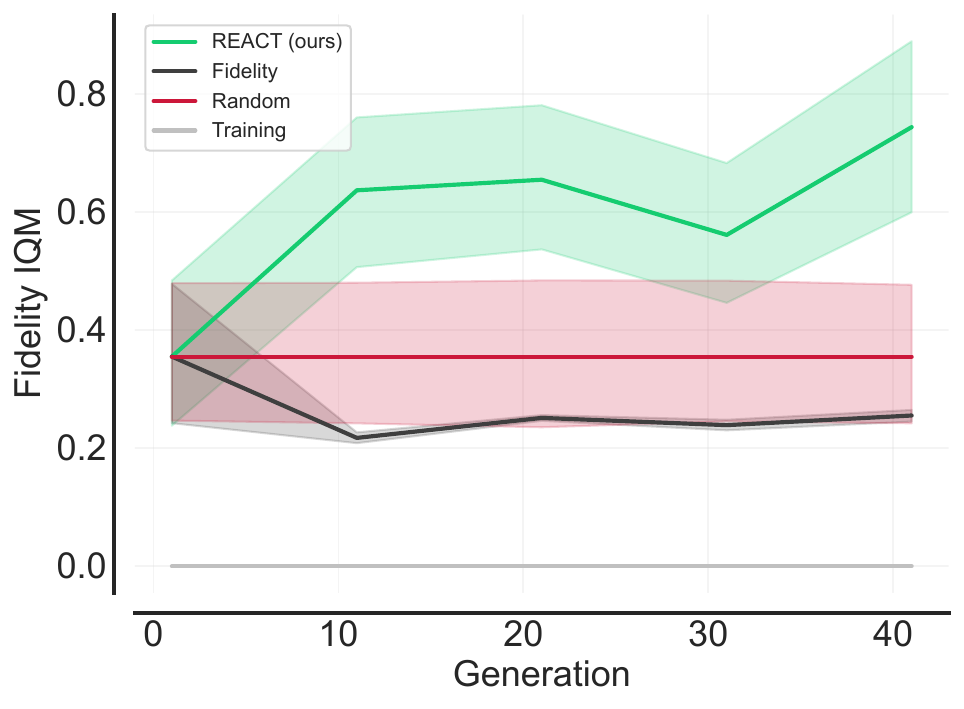}}
\subfloat[Fitness Composition\label{fig:FlatGrid-Fitness}]{\includegraphics[width=0.32\linewidth]{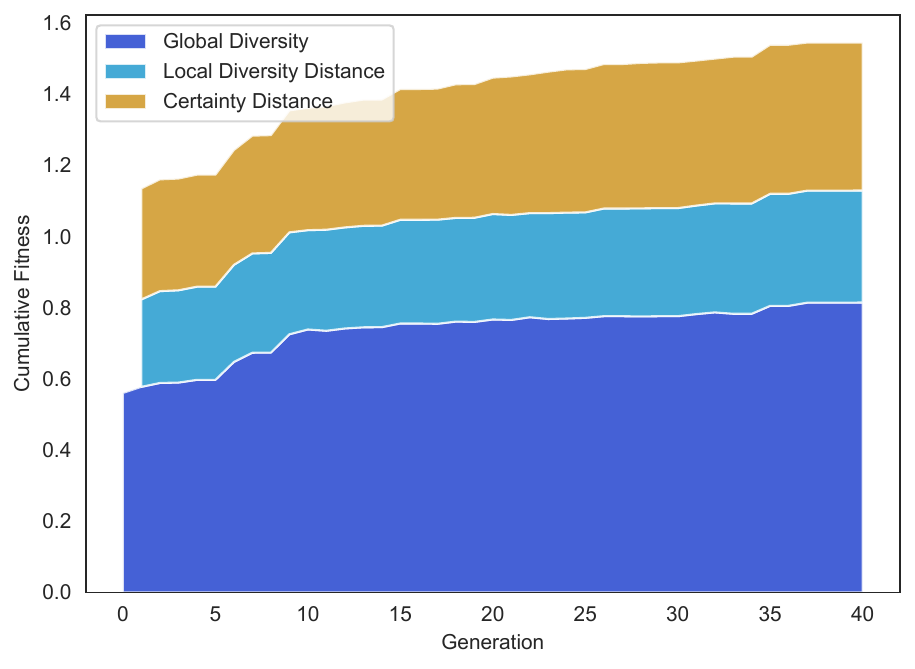}}\\
\subfloat[Ablation Study of Fitness Components\label{fig:FlatGrid-Ablations}]{\includegraphics[width=0.55\linewidth]{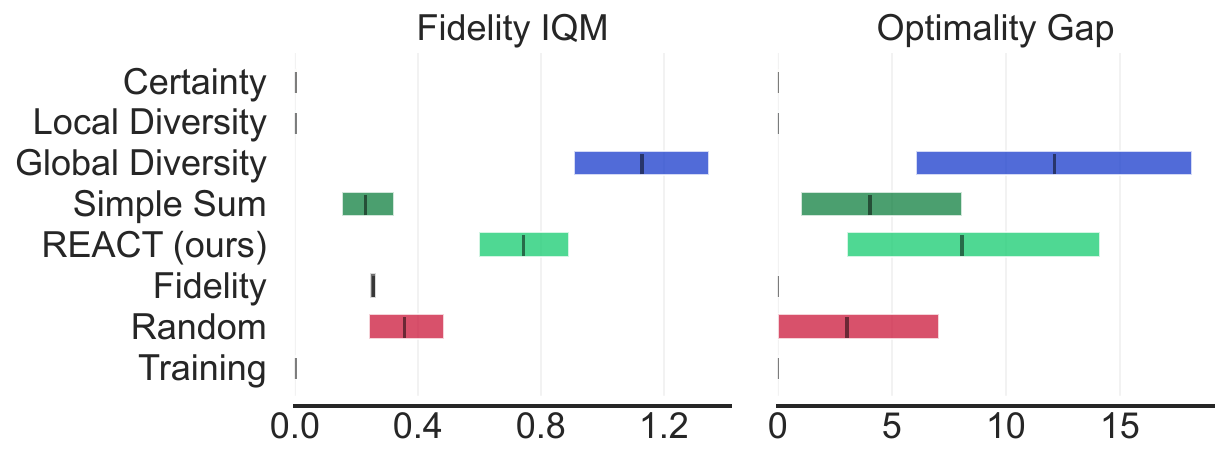}}
\subfloat[REACT (ours)\label{fig:REACTHM-FlatGrid}]{\raisebox{0.25cm}{\includegraphics[width=0.15\linewidth]{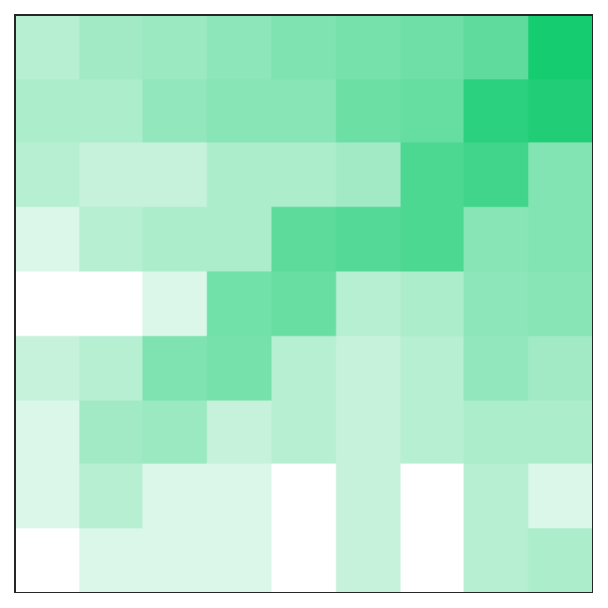}}}
\subfloat[Fidelity\label{fig:FidelityHM-FlatGrid}]{\raisebox{0.25cm}{\includegraphics[width=0.15\linewidth]{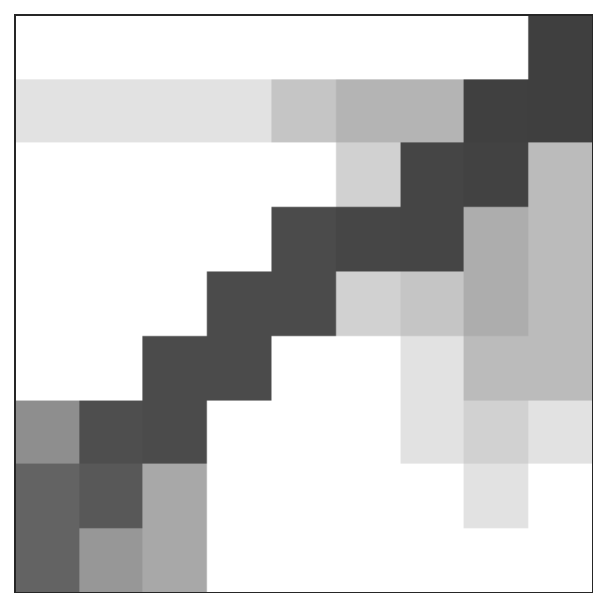}}}
\subfloat[Random\label{fig:RandomHM-FlatGrid}]{\raisebox{0.25cm}{\includegraphics[width=0.15\linewidth]{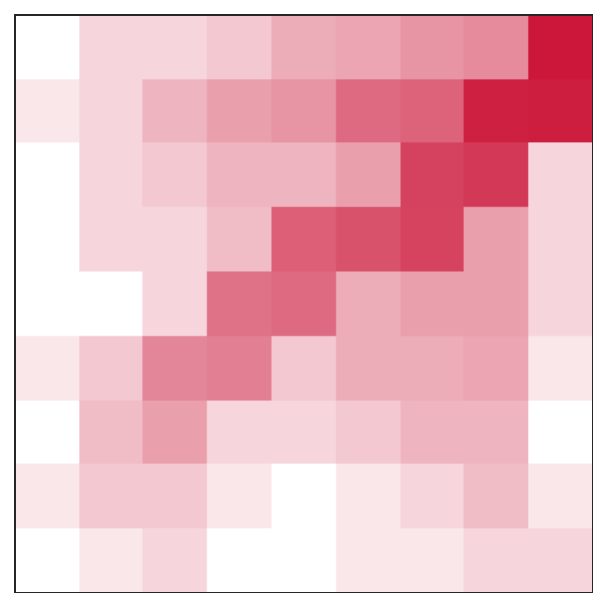}}}
\caption{Evaluation of REACT-generated policy demonstrations in the \texttt{FlatGrid11} \protect\subref{fig:FlatGrid11} w.r.t. the fidelity optimization progress \protect\subref{fig:FlatGrid-Fidelity}, fitness composition \protect\subref{fig:FlatGrid-Fitness}, and final fidelity IQMs and return optimality gaps \protect\subref{fig:FlatGrid-Ablations}, comparing REACT with joint fitness (light green), a simple sum (dark green), certainty (orange), local- (light blue), and global diversity (dark blue) fitness to random initial states (red), training initial states (light grey), fidelity-based optimization (dark grey), and visualizations of the states visited by REACT \protect\subref{fig:REACTHM-FlatGrid}, Fidelity \protect\subref{fig:FidelityHM-FlatGrid} and Random-generated \protect\subref{fig:RandomHM-FlatGrid} demonstrations.}\label{fig:Eval-FlatGrid}
\end{figure*}

To assess the effectiveness of REACT in optimizing diverse and interpretable policy demonstrations, we compare the fidelity of generated trajectories against multiple baselines, including the training initial state, randomly selected initial states, and an optimization using fidelity as the direct fitness function.
Evaluation results in the \texttt{FlatGrid11} environment (cf. Fig.~\ref{fig:FlatGrid11}) are depicted in Fig.~\ref{fig:Eval-FlatGrid}.
\\[4pt]
Fig.~\ref{fig:FlatGrid-Fidelity} shows the development of demonstration fidelity over 40 optimization generations. 
Using the single initial training state yields a fidelity score of 0. 
Introducing \textit{random} initial states improves this fidelity score to around $0.35$, indicating that simple state diversification already enhances demonstration quality. 
Further optimizing these states using \textit{fidelity} as the direct optimization target, however, decreases the fidelity to approximately $0.25$. 
Yet, REACT significantly outperforms all baselines, achieving a fidelity of around $0.75$ after 40 generations. This demonstrates that optimizing for diversity, certainty, and behavioral novelty rather than directly maximizing fidelity yields superior trajectory selection for interpretability.
\\[4pt]
To analyze the role of each fitness component in optimizing interpretable demonstrations, Fig.~\ref{fig:FlatGrid-Fitness} provides a stacked plot illustrating the composition of the joint fitness function throughout optimization. 
Specifically, it tracks the contributions of global diversity (Eq.~\ref{eq:global_diversity}) and the minimum distances (Eq.~\ref{eq:min-local}) of local diversity (Eq.~\ref{eq:local_diversity}), and certainty (Eq.~\ref{eq:certainty}). 
Overall, the results demonstrate that all components contribute meaningfully to the overall fitness and that the joint fitness function serves as a suitable surrogate optimization target, steadily improving over time.
\\[4pt]
To further validate the effectiveness of the joint fitness function, we conduct an ablation study comparing REACT's full fitness function to variations using only certainty (Eq.~\ref{eq:certainty}), local diversity (Eq.~\ref{eq:local_diversity}), or global diversity (Eq.~\ref{eq:global_diversity}). Additionally, we evaluate a simple sum of the three sub-metrics (without applying the minimum distance calculation for local components) and contrast these with using fidelity directly as a fitness function, random initial states, and the training initial state.
Fig.~\ref{fig:FlatGrid-Ablations} compares these approaches using the final fidelity interquartile mean (IQM) (higher values preferred) and the cumulative reward optimality gap (where a high range and mean indicate high diversity and significant deviation from optimal behavior). 
The results confirm that the proposed fitness metrics yield the highest final fidelities, with REACT performing only slightly worse than optimizing for global diversity alone, which may be due to the simplicity of the environment. 
In contrast, random, unoptimized initial states result in significantly lower fidelity scores, highlighting the necessity of trajectory optimization for policy interpretability.
Using fidelity directly as a fitness function performs worse than even a straightforward sum of the fitness components, reinforcing the idea that optimizing for diversity and uncertainty is more effective than directly matching trajectory behavior. 
Additionally, using only local diversity metrics produces demonstrations similar to those generated from the training state, with fidelity values near 0. 
Looking at the optimality gaps further support these findings: 
Optimizing for global diversity alone yields the largest range and highest values, while the REACT fitness function ranks second, demonstrating significant improvements over random or training-state-based initializations.
\\[4pt]
To qualitatively analyze the distribution of visited states in generated demonstrations, Fig.~\ref{fig:RandomHM-FlatGrid} visualizes all states encountered when starting from random initial states, while Fig.~\ref{fig:REACTHM-FlatGrid} shows the states visited when using REACT-optimized initial states. The results indicate that REACT-generated demonstrations exhibit a higher density of outer-state visits compared to those generated from random initial states, while fidelity optimized trajectories (cf. Fig.~\ref{fig:FidelityHM-FlatGrid}) show even less diverse. This suggests that REACT enables more exploratory and diverse behavior within the environment. However, since the environment remains relatively simple and obstacle-free, the difference between the two distributions is moderate.

\subsection{Holey Gridworlds}

\begin{figure*}[t]\centering
\subfloat[\texttt{HoleyGrid11}~\cite{Altmann_hyphi_gym}\label{fig:HoleyGrid11}]{\includegraphics[width=0.32\linewidth]{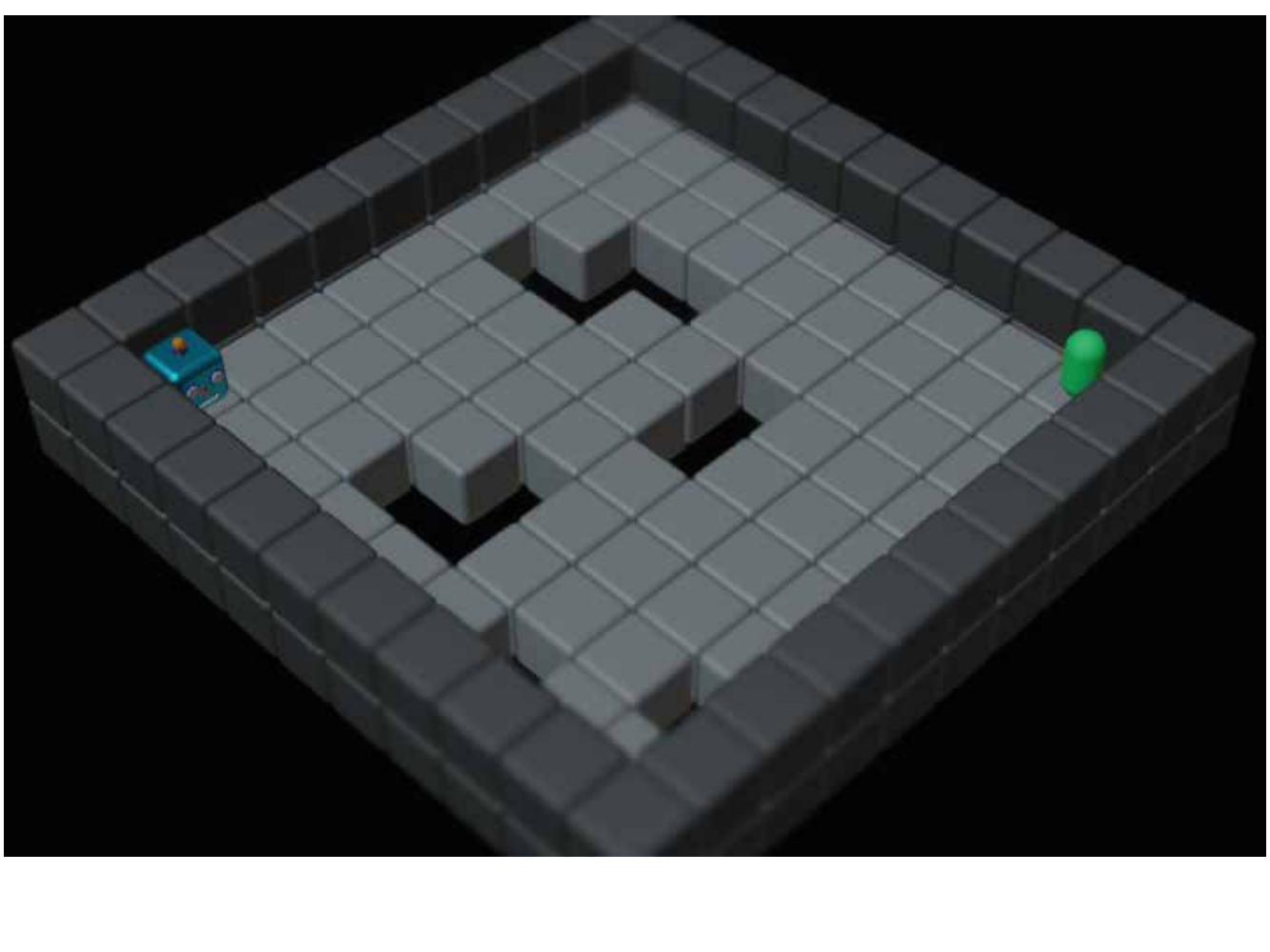}}
\subfloat[Fidelity Optimization\label{fig:HoleyGrid-Fidelity}]{\includegraphics[width=0.32\linewidth]{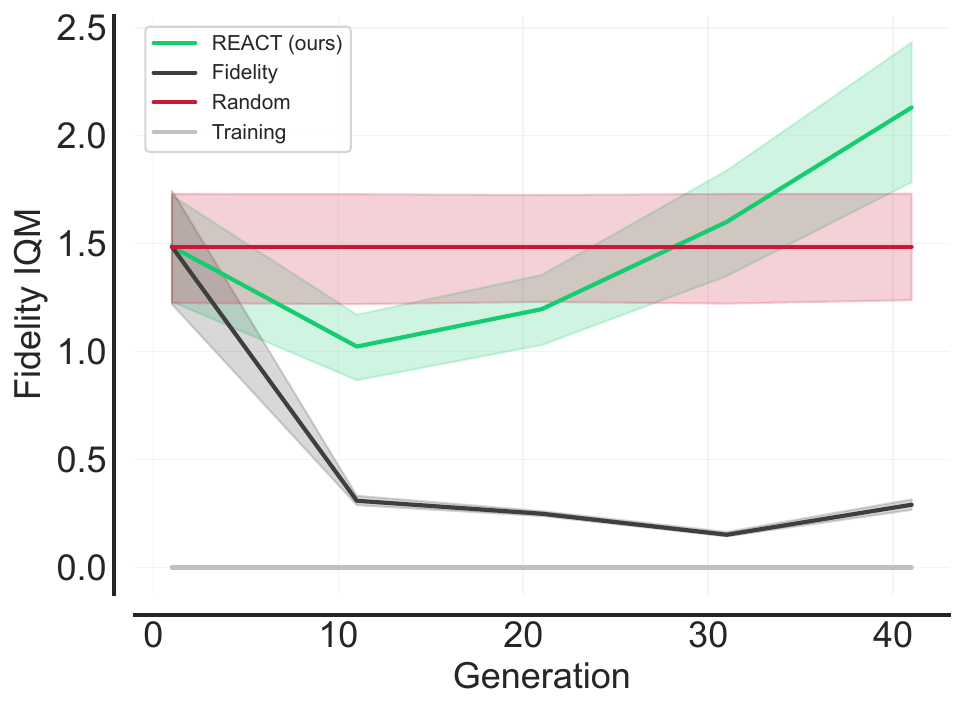}}
\subfloat[Fitness Composition\label{fig:HoleyGrid-Fitness}]{\includegraphics[width=0.32\linewidth]{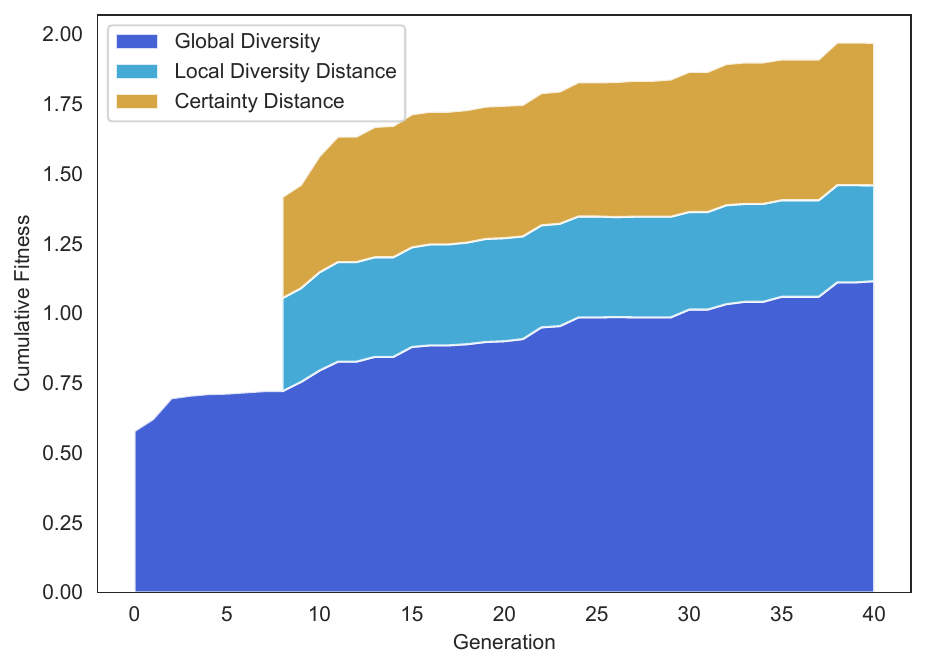}}\\
\subfloat[Ablation Study of Fitness Components\label{fig:HoleyGrid-Ablations}]{\includegraphics[width=0.55\linewidth]{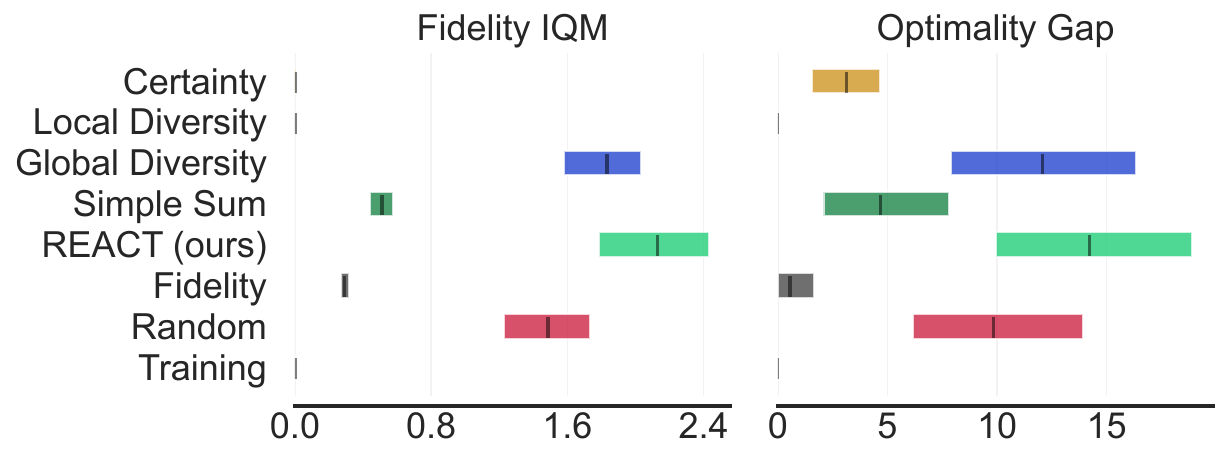}}
\subfloat[REACT (ours)\label{fig:REACTHM-HoleyGrid}]{\raisebox{0.25cm}{\includegraphics[width=0.15\linewidth]{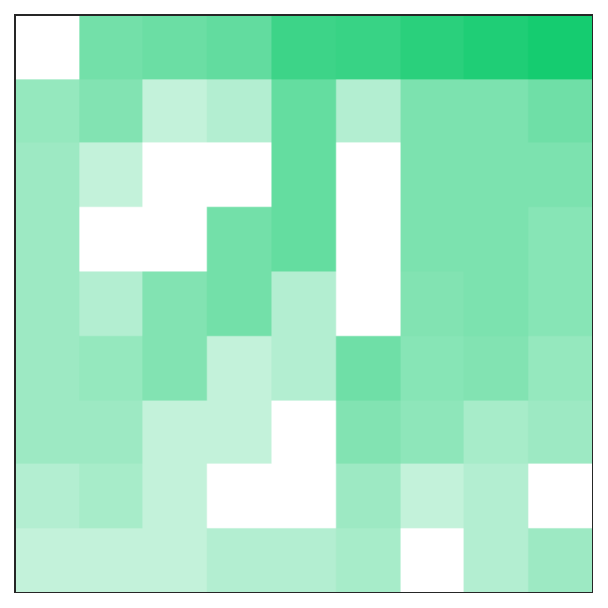}}}
\subfloat[Fidelity\label{fig:FidelityHM-HoleyGrid}]{\raisebox{0.25cm}{\includegraphics[width=0.15\linewidth]{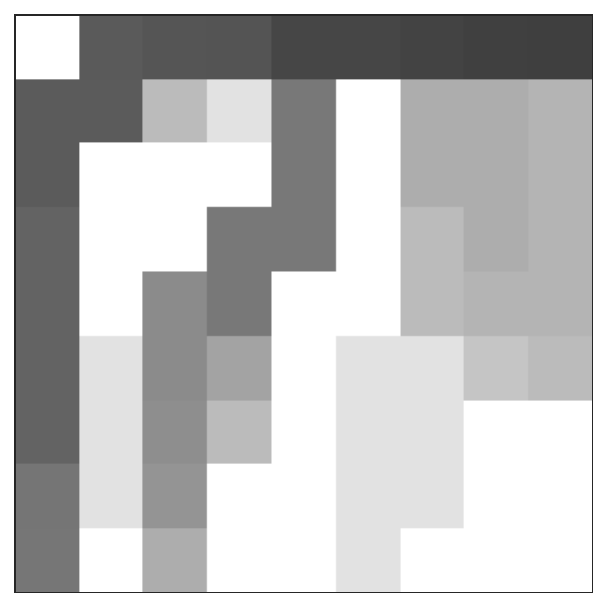}}}
\subfloat[Random \label{fig:RandomHM-HoleyGrid}]{\raisebox{0.25cm}{\includegraphics[width=0.15\linewidth]{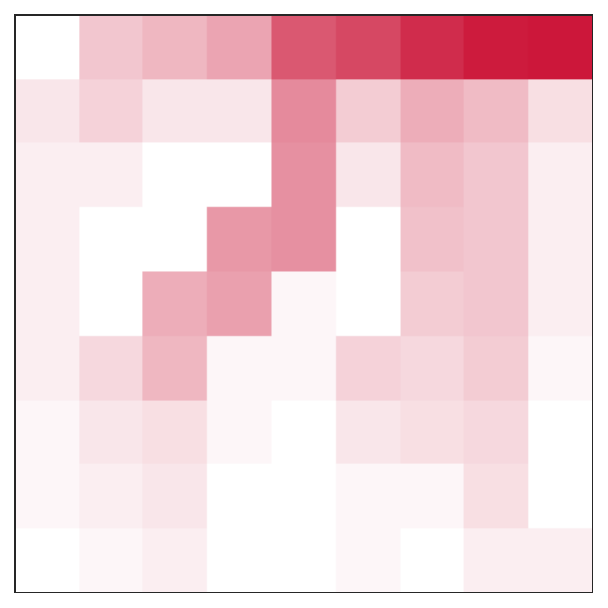}}}
\caption{Evaluation of REACT-generated policy demonstrations in the \texttt{HoleyGrid11} \protect\subref{fig:HoleyGrid11} w.r.t. the fidelity optimization progress \protect\subref{fig:FlatGrid-Fidelity}, fitness composition \protect\subref{fig:HoleyGrid-Fitness}, and final fidelity IQMs and return optimality gaps \protect\subref{fig:HoleyGrid-Ablations}, comparing REACT with joint fitness (light green), a simple sum (dark green), certainty (orange), local- (light blue), and global diversity (dark blue) fitness to random initial states (red), training initial states (light grey), fidelity-based optimization (dark grey), and visualizations of the states visited by REACT \protect\subref{fig:REACTHM-HoleyGrid}, Fidelity \protect\subref{fig:FidelityHM-HoleyGrid}, and Random \protect\subref{fig:RandomHM-HoleyGrid} demonstrations.}\label{fig:Eval-HoleyGrid}
\end{figure*}

To further assess the effectiveness of REACT, we evaluate its performance in the more complex \textit{HoleyGrid} environment, shown in Fig.~\ref{fig:HoleyGrid11}. This environment extends the previous \textit{FlatGrid} by introducing holes that immediately terminate an episode with a reward of $-50$. These obstacles increase the complexity of the environment, requiring the policy to carefully navigate around them to reach the goal successfully. The analyzed policy was trained using PPO for 150k steps in a static environment, just reaching successful behavior with an average return of $36$ and a trajectory length of $14$.  
\\[4pt]
Fig.~\ref{fig:HoleyGrid-Fidelity} illustrates the development of demonstration fidelity over 40 optimization generations, comparing REACT to several baseline approaches. The optimal training trajectory, which starts from a fixed initial state, achieves a fidelity score of $0$, indicating that it does not sufficiently cover diverse behaviors. Introducing random initial states improves the fidelity to approximately $1.5$, demonstrating that even naive diversification of initial conditions enhances demonstration quality. However, directly optimizing for fidelity as the fitness function again leads to a decline in fidelity, with values dropping below $0.3$ throughout the optimization process. This suggests that directly maximizing fidelity does not necessarily yield the most informative demonstrations, likely due to the lack of diversity considerations.  
In contrast, REACT initially experiences a slight decrease in fidelity within the first ten generations, but this trend reverses as optimization progresses. After generation 10, the fidelity steadily improves, ultimately reaching an average above $2$ by the end of the 40 generations. 
\\[4pt]
This turning point coincides with the fitness composition analysis shown in Fig.~\ref{fig:HoleyGrid-Fitness}, which provides insight into the contributions of different fitness components throughout the optimization process. Initially, global diversity dominates the fitness landscape, driving the search for diverse starting conditions. However, after generation 10, local diversity and certainty become increasingly relevant, refining the selection of demonstrations to highlight meaningful variations in policy behavior. The overall increase in the joint fitness function over time again confirms that REACT provides an effective surrogate optimization target, steadily improving trajectory interpretability.  
\\[4pt]
To further investigate the role of each fitness component, we conduct an ablation study, comparing REACT’s joint fitness function to several alternative formulations. The results, shown in Fig.~\ref{fig:HoleyGrid-Ablations}, evaluate the final demonstration fidelity and the cumulative reward optimality gap across different fitness definitions. Using only local diversity or certainty again fails to provide a meaningful optimization target, yielding fidelity values close to 0. Meanwhile, relying solely on global diversity performs slightly worse than the full joint fitness function, reinforcing the importance of integrating local behavioral variations and uncertainty considerations. The simple sum of the three sub-metrics also proves suboptimal, validating REACT’s explicit formulation of local fitness through the minimum-distance approach.  
Overall, REACT significantly outperforms the compared baselines in this slightly more complex environment. 
The optimality gap analysis further supports these findings. REACT’s joint fitness function achieves both the highest mean and the broadest range of reward deviations, highlighting its ability to generate diverse yet informative demonstrations. While random initial states and the global diversity-only approach perform slightly worse, they still outperform the remaining baselines. In contrast, fitness formulations based solely on local diversity or certainty fail to yield sufficiently diverse demonstrations. Notably, the training initial state results in an optimality gap of 0, which might misleadingly suggest optimal behavior. This underscores the necessity of optimizing diverse trajectories to fully understand the policy beyond its default execution.  
\\[4pt]
Finally, we analyze the state coverage of generated demonstrations by comparing random initial states (Fig.~\ref{fig:RandomHM-HoleyGrid}) to REACT-optimized initial states (Fig.~\ref{fig:REACTHM-HoleyGrid}). The results show that REACT-generated demonstrations exhibit greater coverage of the state space, particularly in the outlying regions of the environment. This confirms that REACT not only optimizes for diverse starting conditions but also encourages exploration of challenging and previously underrepresented states. It is important to note that the white areas in these figures represent holes, which cannot be visited, further emphasizing the importance of generating trajectories that successfully navigate around obstacles.  
\\[4pt]
Overall, the results demonstrate that REACT significantly improves the interpretability of learned policies by optimizing for diverse, high-fidelity trajectories. Compared to baseline methods, REACT consistently produces demonstrations that better capture the variability in policy behavior while ensuring that these demonstrations remain meaningful and informative. By balancing global diversity, local variations, and certainty, REACT enables a structured evaluation of RL policies that goes beyond traditional performance metrics, making it a valuable tool for policy analysis and explainability.

\subsection{Continous Robotic Control}

\begin{figure*}[htb]\centering
\subfloat[\texttt{FetchReach} \cite{Altmann_hyphi_gym}\label{fig:FetchReach-Env}]{\includegraphics[width=0.2\linewidth]{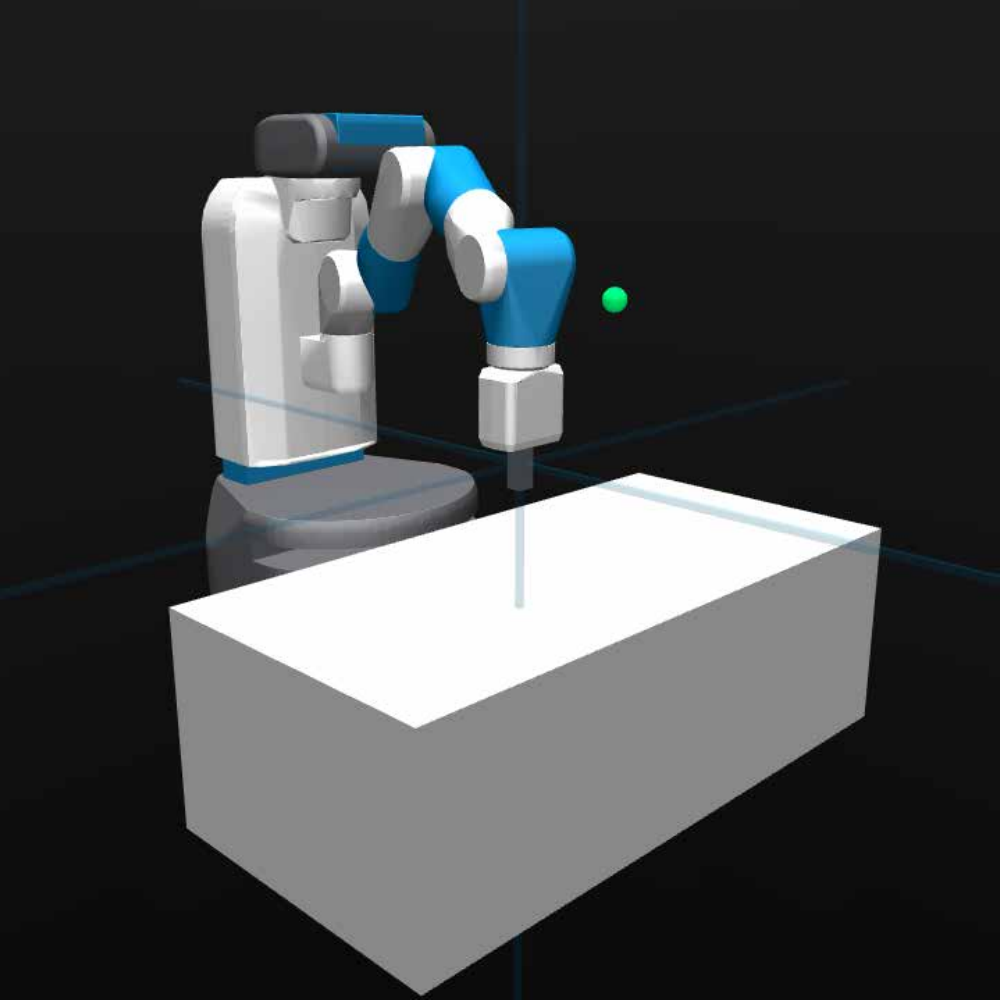}}
\subfloat[Training Return\label{fig:FetchReach-Training}]{\includegraphics[width=0.2\linewidth]{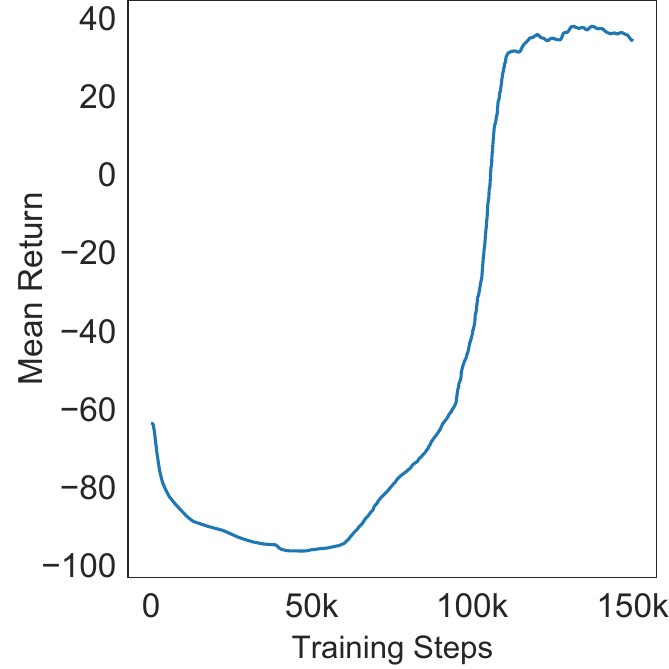}}
\subfloat[Trajectories (50k)\label{fig:Fetch50k-Trajectories}]{\includegraphics[width=0.2\linewidth]{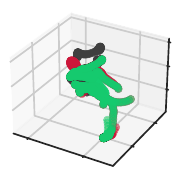}}
\subfloat[Trajectories (100k)\label{fig:Fetch100k-Trajectories}]{\includegraphics[width=0.2\linewidth]{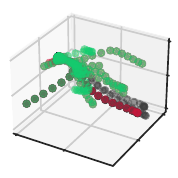}}
\subfloat[Trajectories (150k)\label{fig:Fetch150k-Trajectories}]{\includegraphics[width=0.2\linewidth]{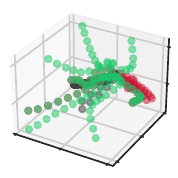}}\\
\subfloat[Fidelity (50k)\label{fig:Fetch50k-Fidelity}]{\includegraphics[width=0.25\linewidth]{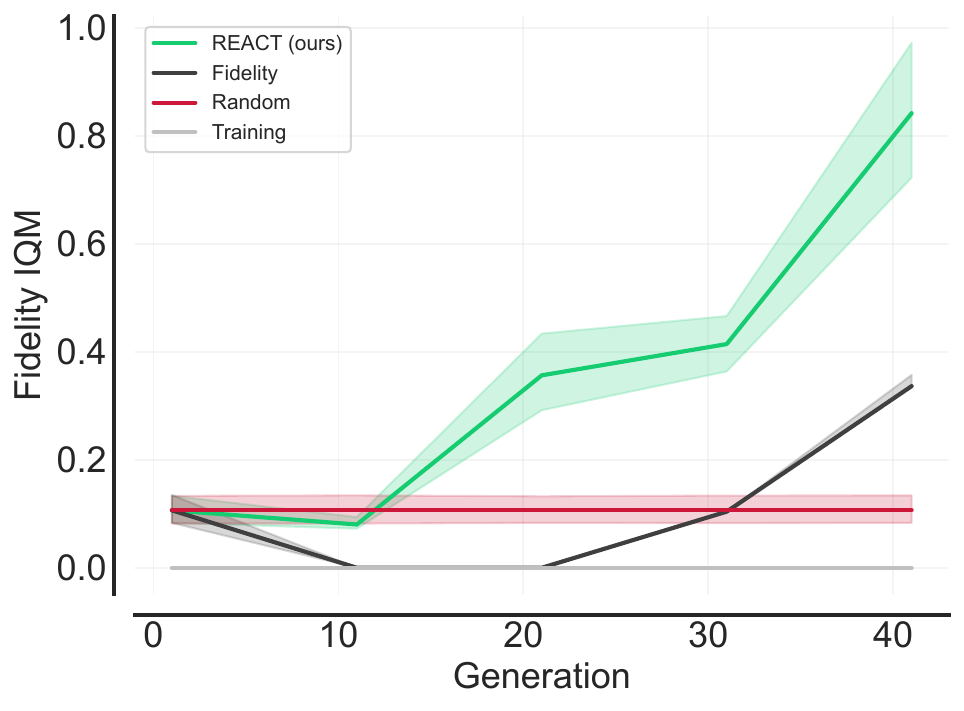}}
\subfloat[Fitness Composition (50k)\label{fig:Fetch50k-Fitness}]{\includegraphics[width=0.25\linewidth]{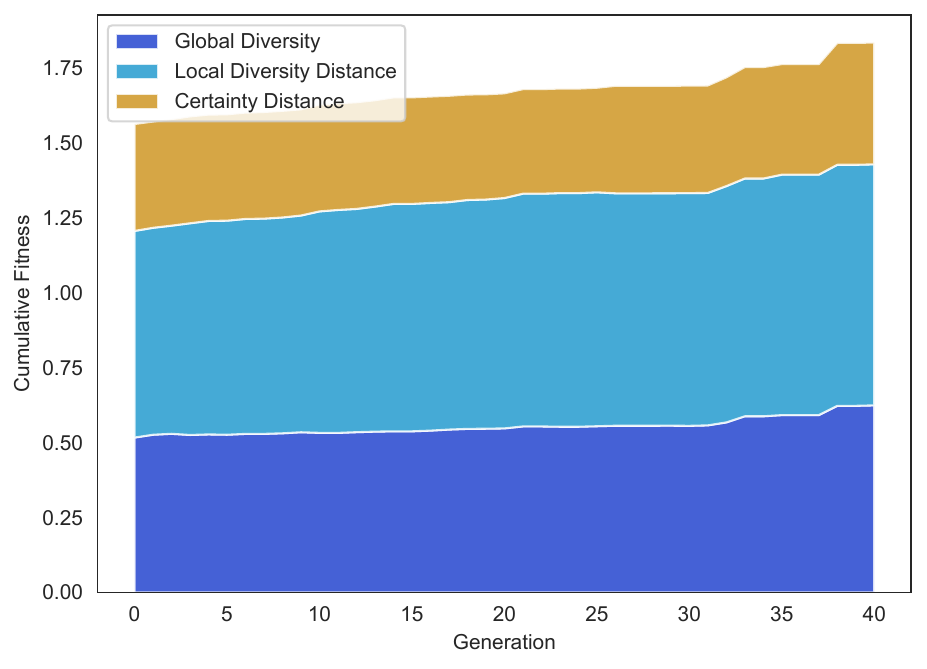}}
\subfloat[Ablations (50k)\label{fig:Ablations-Fetch50k}]{\includegraphics[width=0.5\linewidth]{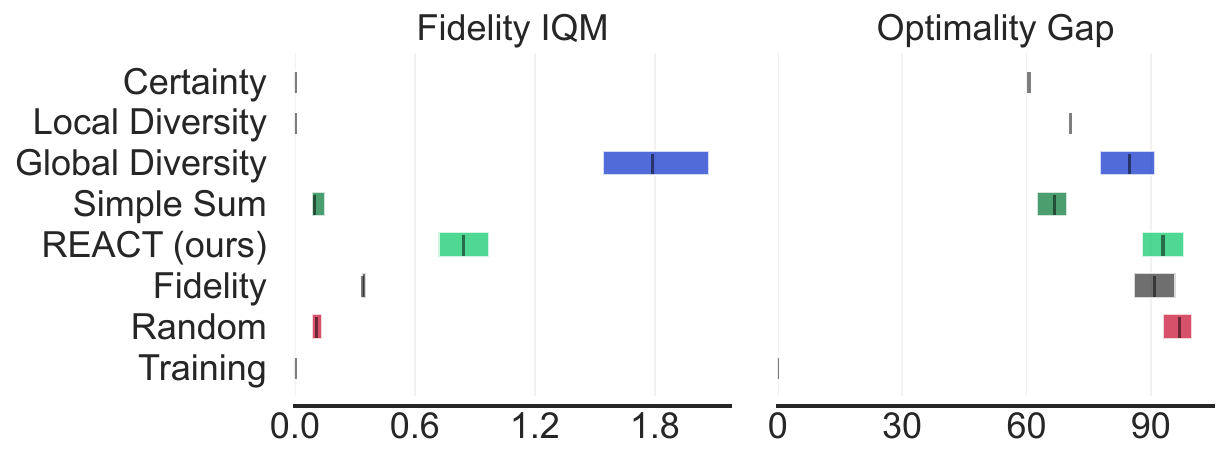}}\\
\subfloat[Fidelity (100k)\label{fig:Fetch100k-Fidelity}]{\includegraphics[width=0.25\linewidth]{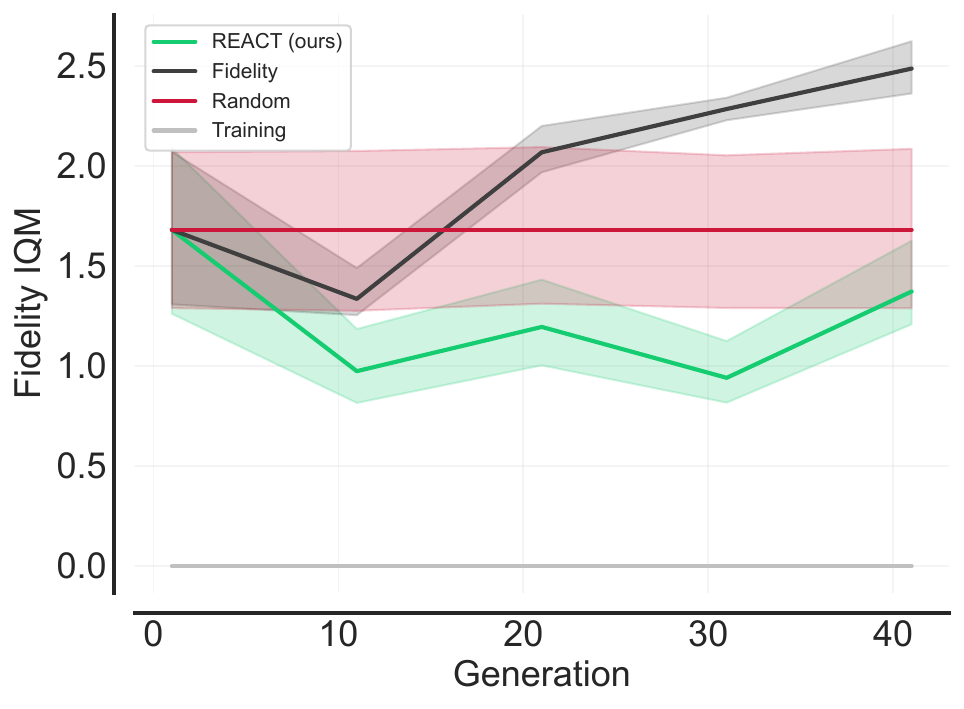}}
\subfloat[Fitness Composition (100k)\label{fig:Fetch100k-Fitness}]{\includegraphics[width=0.25\linewidth]{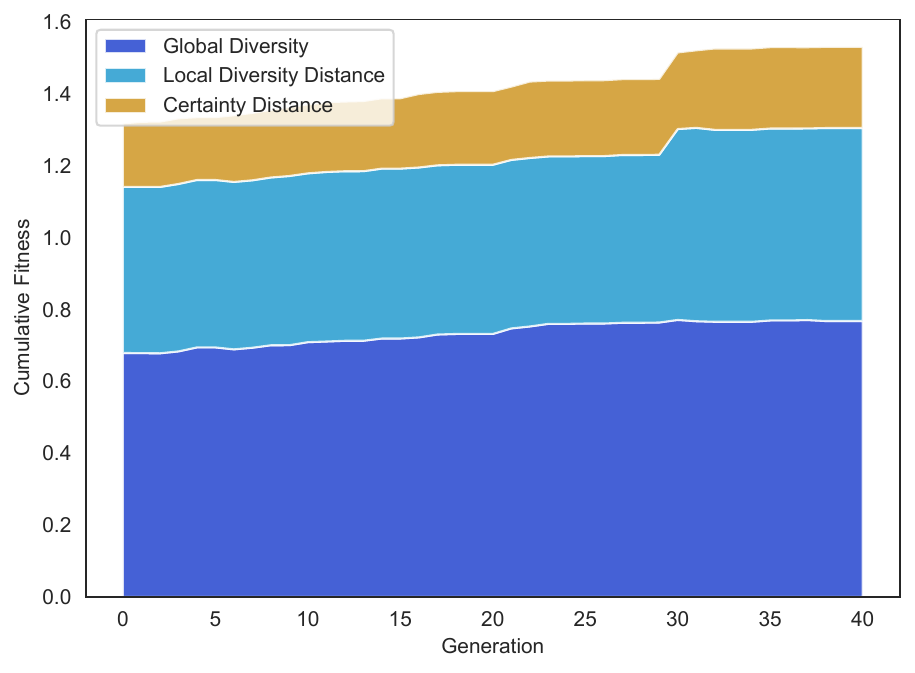}}
\subfloat[Ablations (100k)\label{fig:Ablations-Fetch100k}]{\includegraphics[width=0.5\linewidth]{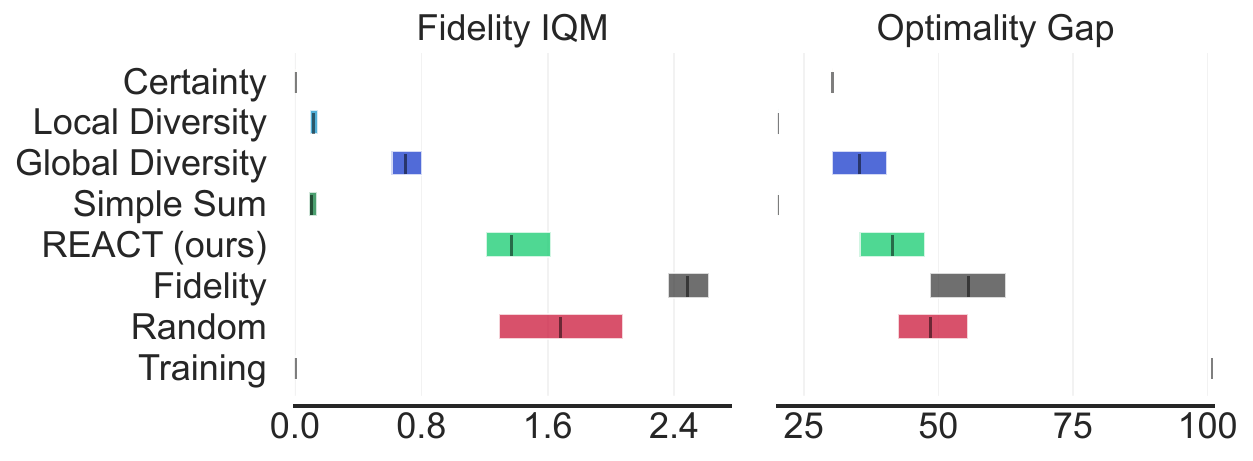}}\\
\subfloat[Fidelity (150k)\label{fig:Fetch150k-Fidelity}]{\includegraphics[width=0.25\linewidth]{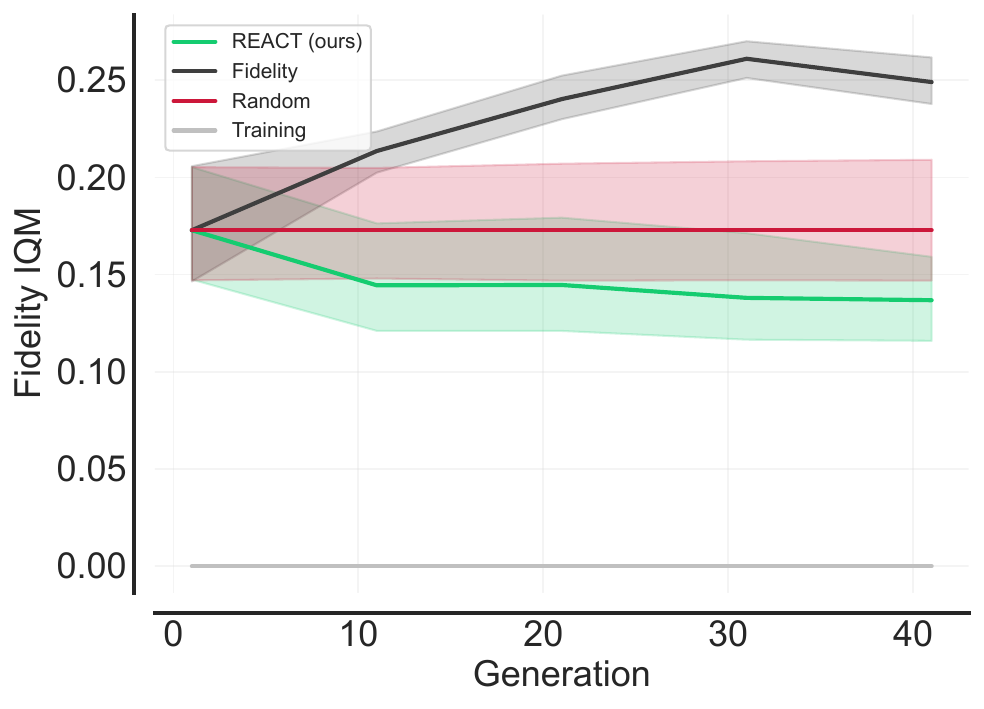}}
\subfloat[Fitness Composition (150k)\label{fig:Fetch150k-Fitness}]{\includegraphics[width=0.25\linewidth]{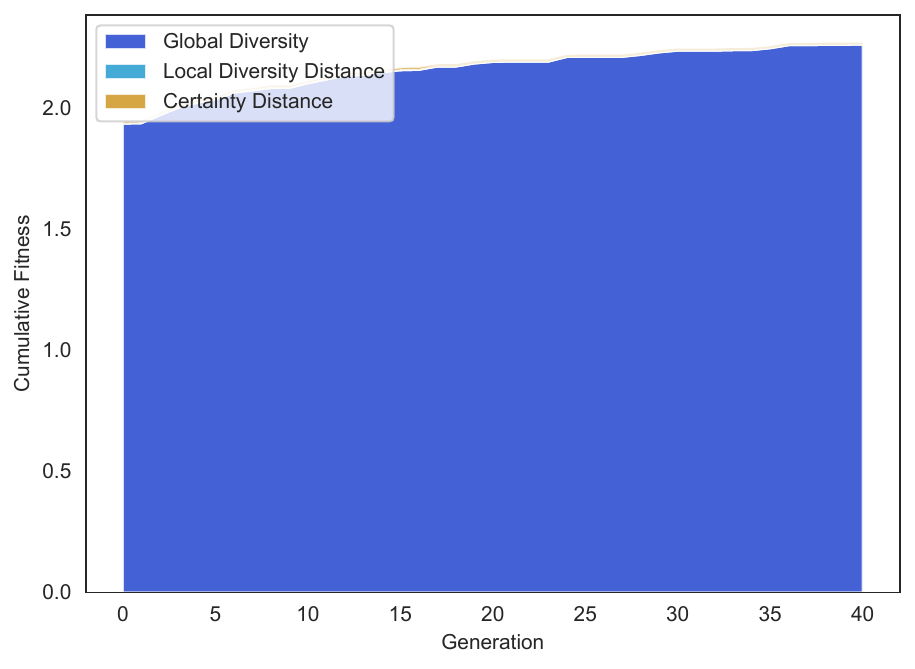}}
\subfloat[Ablations (150k)\label{fig:Ablations-Fetch150k}]{\includegraphics[width=0.5\linewidth]{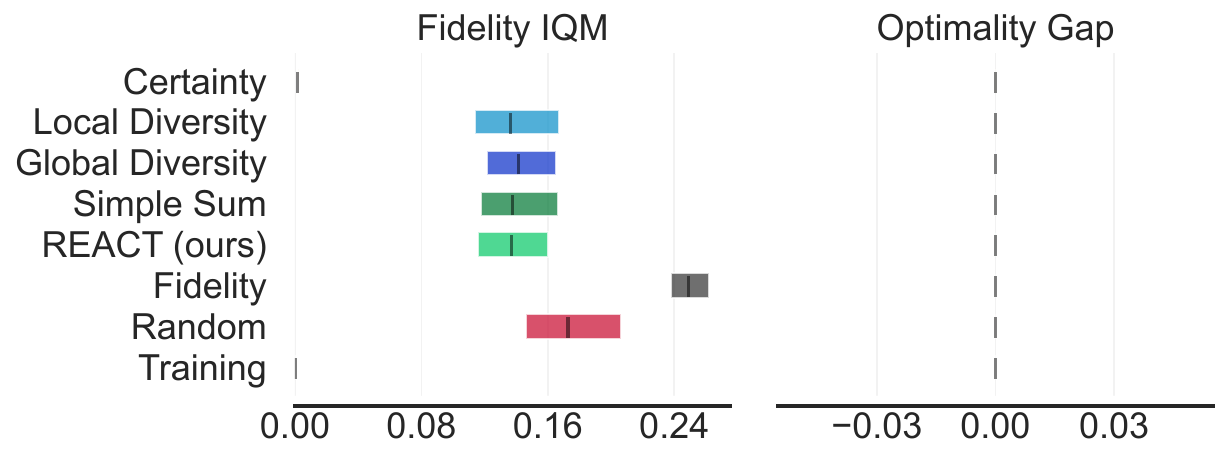}}
\caption{Evaluation of REACT-generated policy demonstrations in the \texttt{FetchReach} task \protect\subref{fig:FetchReach-Env} at three training stages: SAC-50k (\protect\subref*{fig:Fetch50k-Fidelity} - \protect\subref*{fig:Ablations-Fetch50k}), SAC-100k (\protect\subref*{fig:Fetch100k-Fidelity} - \protect\subref*{fig:Ablations-Fetch100k}), and SAC-150k (\protect\subref*{fig:Fetch150k-Fidelity} - \protect\subref*{fig:Ablations-Fetch150k})
 w.r.t. the fidelity optimization progress (\protect\subref*{fig:Fetch50k-Fidelity},~\protect\subref*{fig:Fetch100k-Fidelity},~\protect\subref*{fig:Fetch150k-Fidelity}), joint fitness composition (\protect\subref*{fig:Fetch50k-Fitness},~\protect\subref*{fig:Fetch100k-Fitness},~\protect\subref*{fig:Fetch150k-Fitness}), and final fidelity IQMs and return optimality gaps (\protect\subref*{fig:Ablations-Fetch50k},~\protect\subref*{fig:Ablations-Fetch100k},~\protect\subref*{fig:Ablations-Fetch150k}), comparing REACT with joint fitness (light green), a simple sum (dark green), certainty (orange), local- (light blue), and global diversity (dark blue) fitness to random initial states (red), training initial states (light grey), fidelity-based optimization (dark grey), and visualizations of the resulting demonstration trajectories generated by REACT (green), Fidelity (grey), and Random (red) for a SAC policy trained for 50k \protect\subref{fig:Fetch50k-Trajectories}, 100k \protect\subref{fig:Fetch100k-Trajectories}, and 150k \protect\subref{fig:Fetch150k-Trajectories} steps. Training progress of the SAC policy is shown in \protect\subref{fig:FetchReach-Training}. REACT performs best during early training, is competitive at intermediate stages, and converges with other methods as policies mature.}\label{fig:Eval-FetchReach}
\end{figure*}

To assess REACT's effectiveness in more complex, real-world inspired tasks, we evaluate it on the continuous robotic control benchmark \textit{FetchReach}, as shown in Fig.~\ref{fig:FetchReach-Env}. Here, a 6-DoF robotic manipulator must move its gripper to reach a designated target in 3D space. The environment features continuous state and action spaces, with sparse rewards based on proximity to the target. Each episode spans a fixed length of 50 time steps, and a penalty of $-1$ is applied for every time step during which the distance between the gripper and the target exceeds a tolerance of $0.05$. During training, the gripper always starts at the origin \((0,0,0)\), and the target is sampled randomly within a cube of side length 0.5 centered at the origin. This promotes generalization and robustness in the learned policy.
\\[4pt]
For evaluation, we fix the target location to \((0,0,0)\) and vary only the initial gripper position, encoded using 9 bits per dimension, resulting in fine-grained sampling across the 3D workspace. This allows clearer rendering of demonstration trajectories and facilitates the comparative analysis of different policy behaviors. REACT is run with a population size of 10 individuals for 40 generations, consistent with earlier experiments. Instead of cumulative heatmaps, we visualize the trajectories resulting from a single evaluation seed\footnote{Video renderings are available at \url{https://github.com/philippaltmann/REACT}.}. 
\\[4pt]
We analyze three \textit{Soft Actor Critic} (SAC) \cite{SAC} policies trained for 50k, 100k, and 150k steps respectively, representing distinct phases of policy learning: early, intermediate, and converged. 
The results are summarized in Fig.~\ref{fig:Eval-FetchReach}.
Training performance is illustrated in Fig.~\ref{fig:FetchReach-Training}. These policies allow us to assess how the maturity of the learned behavior influences the optimization and interpretability of demonstrations.

\subsubsection*{SAC-50k: Early Learning Phase}

At 50k training steps, the policy is still in an early phase of learning, with generally low reward and inconsistent success in reaching the target. Demonstrations from random initial states result in a low mean fidelity of 0.1 (Fig.~\ref{fig:Fetch50k-Fidelity}). Directly optimizing fidelity increases this to about 0.3, while REACT further improves it to approximately 0.8. The fitness composition in Fig.~\ref{fig:Fetch50k-Fitness} confirms that all components—global diversity, local diversity, and certainty—meaningfully contribute to the joint fitness throughout optimization. Interestingly, as shown in Fig.~\ref{fig:Ablations-Fetch50k}, optimizing solely for global diversity outperforms all other ablations and even REACT itself, likely due to the policy's limited responsiveness to more nuanced local differences at this stage. Optimality gaps remain high across all methods, consistent with the immature nature of the policy. Trajectories visualized in Fig.~\ref{fig:Fetch50k-Trajectories} reveal scattered and coarse movement toward the goal, suggesting partial task understanding. Overall, REACT is effective at highlighting the partial competencies and behavioral inconsistencies of this early-stage policy.

\subsubsection*{SAC-100k: Intermediate Performance}

By 100k steps, the policy has begun reliably reaching the target but still lacks consistent precision and efficiency. Demonstrations from random initial states yield a fidelity of approximately 1.5 (Fig.~\ref{fig:Fetch100k-Fidelity}). REACT initially reduces fidelity but recovers it to near baseline by the end of optimization. In contrast, using fidelity directly as a fitness function proves much more effective here, boosting the fidelity to around 2.5. This shift suggests that in continuous control environments, particularly with intermediate policies, direct optimization of fidelity may serve as a more suitable surrogate objective than diversity-based criteria. Fig.~\ref{fig:Fetch100k-Fitness} shows a decreasing influence of local diversity and certainty, consistent with the policy’s increasing stability. However, all components remain relevant throughout optimization.
Ablation results (Fig.~\ref{fig:Ablations-Fetch100k}) show that REACT’s joint fitness significantly outperforms variants using only partial components, yet it is outperformed by fidelity-based optimization in this setting. Despite differences in fidelity, the actual demonstrations in Fig.~\ref{fig:Fetch100k-Trajectories} show similarly diverse coverage of the workspace for both REACT and fidelity-based methods, suggesting that REACT remains competitive in terms of qualitative trajectory spread, even when fidelity metrics lag slightly behind.

\subsubsection*{SAC-150k: Converged Policy}

At 150k training steps, the SAC policy exhibits mature and stable behavior, consistently succeeding in the task from a wide range of initial positions. As shown in Fig.~\ref{fig:Fetch150k-Fidelity}, this behavioral consistency results in a notably low initial fidelity from random initializations, around 0.17, substantially lower than at earlier training stages. This suggests that the policy’s high competence and confidence reduce behavioral variability, making it harder to induce distinct demonstrations through simple perturbations. REACT, when optimizing for joint fitness, is unable to improve upon this baseline, converging to a similar fidelity. In contrast, directly optimizing for fidelity yields some improvement, reaching final scores of around 0.25. However, these scores remain relatively low overall, confirming that mature policies tend to produce highly consistent behavior, with little deviation across initial conditions.
The fitness composition analysis in Fig.~\ref{fig:Fetch150k-Fitness} further supports this interpretation. The influence of certainty and local diversity declines noticeably, reflecting the high certainty and low variance in the policy’s responses. This is reinforced by the ablation study in Fig.~\ref{fig:Ablations-Fetch150k}, where the certainty component remains near zero throughout optimization. Across all ablated variants, fidelity remains low and relatively uniform, with even the random initialization slightly outperforming the optimized variants — except for the fidelity-optimized demonstrations, which clearly outperform all others in terms of fidelity score.
\\[4pt]
Despite this, the trajectories shown in Fig.~\ref{fig:Fetch150k-Trajectories} highlight a key qualitative distinction: REACT-generated demonstrations span a noticeably wider portion of the 3D space compared to both random and fidelity-optimized approaches. This broader spatial coverage provides a more comprehensive view of how the policy generalizes across different conditions, even when such variations result in similar outcomes. While the resulting fidelity scores are lower, they do not fully capture the added interpretability from seeing how the policy handles extreme or edge-case scenarios. Additionally, the optimality gap for all approaches is effectively zero, indicating that the policy has reached reliable and optimal performance in this environment.
In summary, while REACT offers limited quantitative gains in fidelity at this stage, it still produces qualitatively richer demonstrations that help illustrate the robustness and behavioral consistency of a well-trained policy.

\subsubsection*{Discussion and Summary}
Across all environments and training stages, REACT consistently provides interpretable insights into policy behavior by generating a curated set of diverse demonstrations. In early training phases, where behavior is still unstable or underdeveloped, REACT clearly outperforms random and ablation-based baselines, producing demonstrations with significantly higher fidelity and broader coverage of possible behaviors. For intermediate policies, particularly in continuous control tasks, direct fidelity optimization often yields higher quantitative fidelity scores. However, REACT still identifies behaviorally distinct trajectories that offer qualitative insights, especially in how different regions of the state space are traversed.
These trends are also reflected in Table~\ref{tab:summary}, which summarizes the final fidelity scores across all environments and methods.
\begin{table}[h]\centering
\begin{tabular}{cc|cccc}
Env&Algorithm&  REACT (ours) & Fidelity & Random & Training \\ \hline
FlatGrid11 & PPO-35k   & $\mathbf{0.744\pm0.147}$ & $0.255\pm0.010$ & $0.355\pm0.120$ & $0.000\pm0.000$ \\
HoleyGrid11 & PPO-150k & $\mathbf{2.130\pm0.324}$ & $0.291\pm0.023$ & $1.486\pm0.250$ & $0.000\pm0.000$ \\ \hline
\multirow{3}{*}{FetchReach} & SAC-50k  & $\mathbf{0.842\pm0.127}$ & $0.336\pm0.010$ & $0.106\pm0.025$ & $0.000\pm0.000$ \\
& SAC-100k & $1.373\pm0.212$ & $\mathbf{2.487\pm0.132}$ & $1.681\pm0.392$ & $0.000\pm0.000$ \\
& SAC-150k & $0.137\pm0.022$ & $\mathbf{0.249\pm0.012}$ & $0.173\pm0.030$ & $0.000\pm0.000$  \\ \hline
\end{tabular}\caption{Evaluation Summary: Final Fidelity IQMs}\label{tab:summary}
\end{table}

\noindent Overall, REACT serves as a versatile and policy-agnostic tool for analyzing reinforcement learning agents. It adapts well across different environments and training levels and remains robust to varying degrees of behavioral complexity, highlighting both capabilities and potential blind spots in learned behaviors.

\section{Conclusion}

To enhance the interpretability of reinforcement learning policies, we introduced \textit{Revealing Evolutionary Action Consequence Trajectories} (REACT), a framework for generating diverse and informative behavior demonstrations. By applying structured perturbations to the initial state, REACT induces edge-case behaviors that reveal the true operational characteristics of a learned policy. To select suitable demonstrations, we proposed a joint fitness function that evaluates candidate trajectories by combining their local diversity and certainty with the global diversity across a population of trajectories. This joint measure enables the identification of demonstrations that not only differ from one another but also offer interpretable insight into the structure of the learned behavior.
\\[4pt]
REACT leverages an evolutionary algorithm to optimize this fitness over populations of individuals, where each individual represents an altered initial state. We evaluated REACT on discrete gridworlds (with and without obstacles) and on the continuous robotic control task \textit{FetchReach}, across policies at different training stages. These experiments demonstrated REACT's adaptability and utility across environments and levels of policy maturity.
Our results show that REACT reliably identifies diverse and informative demonstration sets, particularly in the early and intermediate stages of policy training, where behavioral variability is still present. 
In continuous control tasks, REACT helped uncover subtle distinctions in policy behavior and generalization, although direct fidelity optimization occasionally proved more effective in improving quantitative fidelity scores. 
Overall, REACT-generated trajectories exhibited broader spatial coverage and revealed policy characteristics that are not apparent from standard training configurations.
\\[4pt]
Importantly, we analyzed demonstration quality using an adapted fidelity metric, providing a statistical proxy for how well the demonstrations represent the policy's behavior space, rather than using the environmental return. While REACT was shown to be highly effective, particularly in discrete and early-phase environments, our findings also indicate that direct fidelity optimization may be more effective in certain continuous control tasks, suggesting opportunities for refining the fitness function to better suit such domains.
\\[4pt]
Our current instantiation of REACT alters only the initial agent state (and optionally, the goal) to generate demonstrations. Future work could extend this framework to explore a broader range of environment configurations, including physical obstacles, dynamic environments, or task redefinitions, enabling a more thorough interrogation of the policy's behavior under diverse conditions. Additionally, the diverse trajectory pool generated by REACT may be used as input for downstream applications such as curriculum refinement, adversarial retraining, or the construction of causal models for behavior explanation.
\\[4pt]
In summary, REACT provides a principled and scalable approach to policy-centric interpretability in reinforcement learning. By focusing on diverse behavioral manifestations of a given policy, rather than only outcome metrics or reward signals, REACT helps bridge the gap between performance evaluation and actionable understanding of policy behavior. We believe REACT offers a promising foundation for future research on policy inspection, explanation, and robustness in real-world reinforcement learning systems.

\backmatter

\vspace{8pt}\noindent\textbf{Supplementary information. }
All required implementations and resulting video renderings are available at \url{https://github.com/philippaltmann/REACT}.


\bibliography{REACT2}

\end{document}